\documentclass[10pt,journal,compsoc]{IEEEtran}

\usepackage{cite}            
\usepackage{amsmath,amssymb} 
\usepackage{graphicx}        
\usepackage{subfig}          
\usepackage{algorithmic}     
\usepackage{array}           
\usepackage{booktabs}        
\usepackage{url}             
\usepackage{xcolor}          
\usepackage{multirow}
\usepackage{hyperref}
\usepackage[font=small,labelfont=bf]{caption}

\colorlet{cred}{red!40!gray}
\colorlet{cblue}{blue!40!gray}
\colorlet{cyellow}{yellow!70!black}
\colorlet{cgreen}{green!40!gray}
\colorlet{ccyan}{cyan!40!gray}
\colorlet{cpurple}{orange!80!gray}

\newcommand\Rone{\textcolor{cblue}{ \textbf{R1}}}
\newcommand\Rtwo{\textcolor{cyellow}{ \textbf{R2}}}
\newcommand\Rthree{\textcolor{cpurple}{ \textbf{R3}}}
\newcommand\Rfour{\textcolor{cgreen}{ \textbf{R4}}}

\newif\ifshowreviewer
\showreviewerfalse

\newcommand{\reviewtag}[1]{%
  \ifshowreviewer%
    (#1)%
  \fi%
}

\ifshowreviewer
  \colorlet{cred}{red!40!gray}
\else
  \colorlet{cred}{black}
\fi

\begin{document}

\title{50 Years of Automated Face Recognition}


\author{
    Minchul Kim,
    Anil Jain,
    Xiaoming Liu\\
    Department of Computer Science and Engineering,\\
    Michigan State University, East Lansing, MI, 48824\\
    \texttt{\{kimminc2, jain, liuxm\}@cse.msu.edu}
}


\IEEEtitleabstractindextext{%
\setcounter{figure}{0} 

    \captionsetup{type=figure}
    \includegraphics[width=\linewidth]{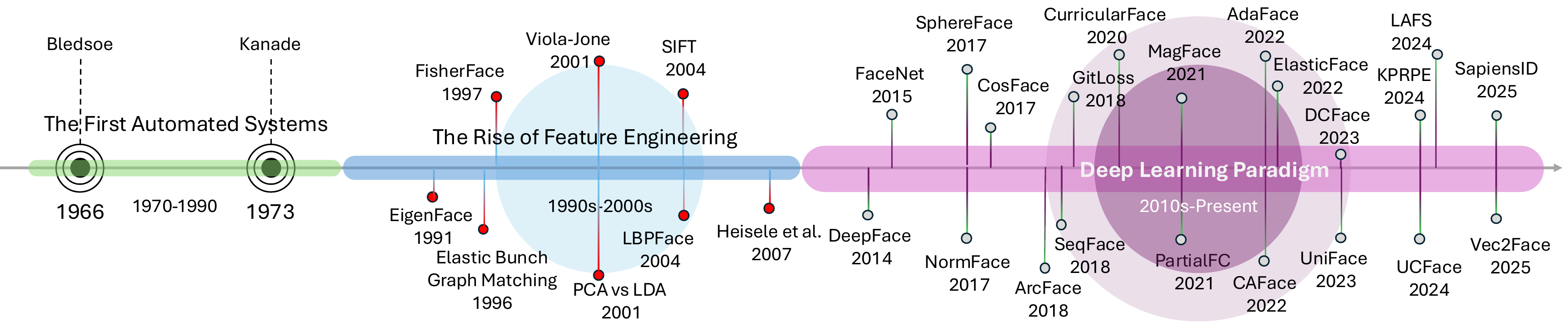}
    \captionof{figure}{\textcolor{cred}{Historical evolution of automated face-recognition research over the past five decades. The timeline illustrates key milestones—from early geometric models (1960s–1980s), through the feature-engineering era (1990s–2000s), to modern deep-learning systems (2010s–present). Colors indicate each era; transparency is for visual purposes only}.\reviewtag{\Rtwo}\\}
    
\begin{abstract} 

Over the past five decades, automated face recognition (FR) has progressed from handcrafted geometric and statistical approaches to advanced deep learning architectures that now approach, and in many cases exceed, human performance. This paper traces the historical and technological evolution of FR, encompassing early algorithmic paradigms through to contemporary neural systems trained on extensive real and synthetically generated datasets. We examine pivotal innovations that have driven this progression, including advances in dataset construction, loss function formulation, network architecture design, and feature fusion strategies. Furthermore, we analyze the relationship between data scale, diversity, and model generalization, highlighting how dataset expansion correlates with benchmark performance gains. \textcolor{cred}{Recent systems have achieved near-perfect large-scale identification accuracy, with the leading algorithm in the latest NIST FRTE 1:N benchmark reporting a FNIR of 0.15 percent at FPIR of 0.001 on a gallery of over 10 million identities} \reviewtag{\Rone} . We delineate key open problems and emerging directions, including scalable training, multi-modal fusion, synthetic data, and interpretable recognition frameworks.

\end{abstract}

}

\maketitle

\IEEEdisplaynontitleabstractindextext
\IEEEpeerreviewmaketitle


\section{Introduction}
\label{sec:introduction}

For half a century, the dream of machines `seeing’ and recognizing faces has captivated researchers and fueled imaginations, leaping from the realm of science fiction to become a pervasive reality. What began as a computationally intractable problem, requiring painstaking manual feature engineering, has blossomed into a cornerstone of modern security, convenience, and even social interaction. However, this rapid ascent has not been without its complexities. The journey from Kanade’s pioneering work~\cite{kanade1974picture} to today’s deep learning behemoths reveals not just a story of algorithmic innovation, but a shifting landscape of ethical considerations, data dependencies, and the ever-present challenge of defining `identity’ itself. This paper chronicles that 50-year evolution, examining the pivotal breakthroughs, the persistent hurdles, and the emerging frontiers that will shape the future of automated face recognition (FR). \textcolor{cred}{Fig.1 presents a timeline summarizing major milestones across five decades of automated 
FR research. \reviewtag{\Rtwo}}

\begin{figure}
    \centering
    \includegraphics[width=1.0\linewidth]{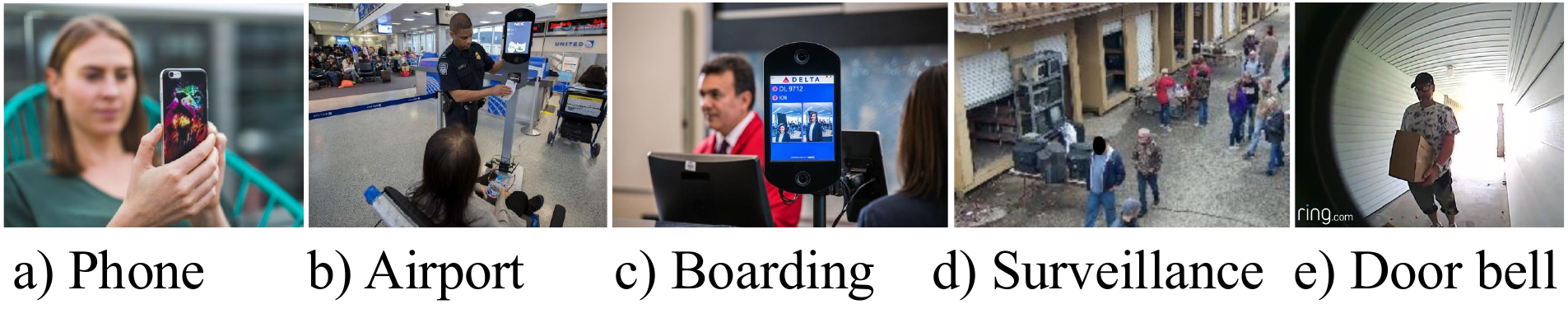}
    \caption{
    Examples of real-world FR applications: 
    \href{https://pixabay.com/photos/people-woman-phone-camera-2572957/}{(a)} cellphone unlocking via facial authentication, 
    \href{https://commons.wikimedia.org/wiki/File:IAH_Houston_Airport_Biometrics_and_CBP_Operations_\%2840077354034\%29.jpg}{(b)} identity verification at airport security checkpoints, 
    \href{https://www.flickr.com/photos/deltanewshub/46092006221/}{(c)} FR for boarding pass verification, 
    \href{https://biometrics.cse.msu.edu/Publications/Face/Kalkaetal_IJBSIARPPAJanusSurveillanceVideoBenchmark_BTAS2018.pdf}{(d)} public surveillance with facial analysis, and 
    \href{https://universe.roboflow.com/finalparcellarge/valdataset-unseen}{(e)} smart doorbells employing FR for home security. 
    These use cases highlight the ubiquity and versatility of FR systems across personal, commercial, and governmental domains.
    }
    \label{fig:fr_application}
    \vspace{-4mm}
\end{figure}

FR has become one of the most prevalent biometric modalities employed today~\cite{afr50}. See Fig.~\ref{fig:fr_application} for representative applications. Several factors contribute to this widespread adoption. Faces can be identified at a distance, offering a non-contact and less intrusive method compared to other biometrics like fingerprints or iris~\cite{clearview}. Face acquisition can be achieved using low-cost cameras, making it accessible and scalable across diverse applications~\cite{clearview,afr50}. The non-contact nature of FR offers hygienic advantages, especially salient in a post-pandemic era~\cite{visage,lai2023post}. Furthermore, FR can be performed covertly using ubiquitous surveillance cameras, and benefits from the existence of extensive legacy databases containing facial images such as passports, visas, mugshots and driver's licenses \cite{afr50}.

\begin{figure}[t]
    \centering
    \includegraphics[width=1.0\linewidth]{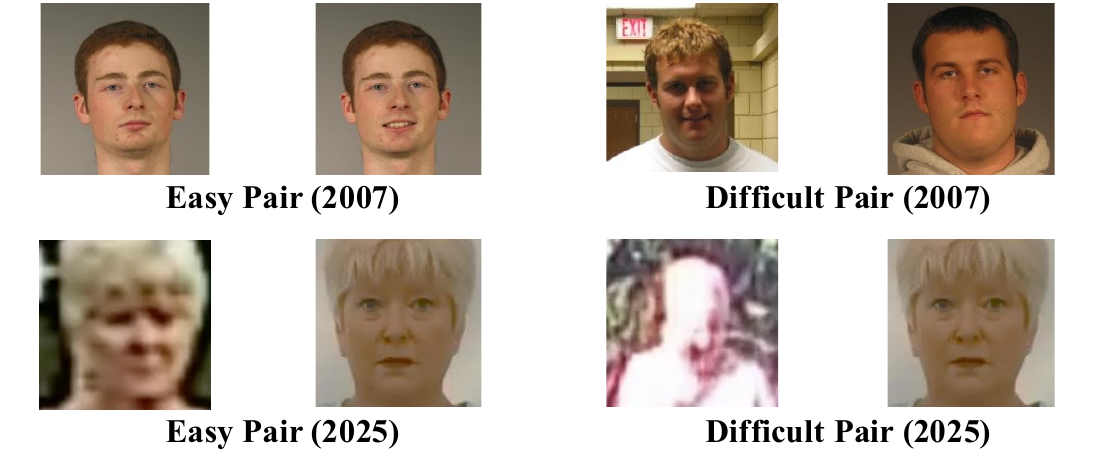}
    \caption{Visualization of easy and difficult face pairs for 2007 (top) and 2025 (bottom) FR, where difficulty is defined by the pairs that (State-of-the-Art) SoTA models of the time struggle to correctly identify~\cite{o2007face,wang2024farsight}. 2007 subjects and images are from ~\cite{o2007face}. 2025 subject and images are from BRIAR dataset~\cite{jager2025expanding}}
    \label{fig:easyhard}
    \vspace{-4mm}
\end{figure}

Notably, even prior to deep mode, automated FR systems demonstrated the potential to surpass human capabilities in certain scenarios. Studies conducted in 2007 indicate that algorithms could outperform average human performance in matching face pairs, particularly in simpler cases \cite{o2007face}. Further research in 2010 revealed that a specific algorithm exceeded the accuracy of thousands of customs inspectors when dealing with straightforward facial comparisons in operational settings \cite{ding2010computers}. Some examples of challenging pairs are given in Fig.~\ref{fig:easyhard}. While these early successes are significant, the field has undergone a dramatic transformation since then, fueled by advancements in deep networks and the availability of large-scale datasets. This paper will revisit this critical juncture, exploring how the field has evolved and the extent to which current FR technology has surpassed human abilities across a wider range of challenging conditions.

The progress in FR has been driven by advances in computing power, the availability of large-scale datasets, and a shift in understanding facial image representation and comparison. Early methods relied on handcrafted features capturing facial structure, including appearance-based approaches like Eigenfaces~\cite{turk1991face} and texture-based ones like Local Binary Patterns (LBP)~\cite{ahonen2004face}. Machine learning methods such as statistical models and cascaded classifiers~\cite{viola2001rapid} improved detection accuracy, enabling more reliable FR pipelines. The advent of deep learning introduced a paradigm shift, allowing algorithms to learn discriminative features directly from large training data, as shown by DeepFace~\cite{taigman2014deepface}, DeepID~\cite{sun2014deep}, and FaceNet~\cite{schroff2015facenet}. These developments greatly enhanced FR performance in accuracy and efficiency but also introduced challenges related to data bias and robustness under pose and occlusion variations.

This paper will trace the historical development of FR, beginning with the foundational work in feature extraction and pattern matching, progressing through the statistical methods that dominated the field for decades, and culminating in the transformative impact of deep learning. We will focus on key innovations in network architecture~\cite{krizhevsky2012imagenet, he2016deep}, loss function design~\cite{wang2017normface, liu2017sphereface, wang2018cosface, deng2019arcface, huang2020curricularface, kim2022adaface}, and the utilization of increasingly large and diverse datasets~\cite{ijbb, ijbc, ijbs, lfw, cfpfp, cplfw, agedb, calfw}. We will address the emerging role of synthetic data as a means to overcome data limitations and mitigate privacy concerns.

Finally, we will discuss the remaining challenges, including adaptation to low-quality images, surpassing human recognition capability, multimodal fusion (such as face and gait), and enhancing the interpretability of complex deep learning models. By providing a comprehensive overview of the field’s past, present, and future, this paper aims to inform both researchers and practitioners and to stimulate further innovation in this field. \textcolor{cred}{Unless otherwise specified, this survey focuses primarily on 2D visible-light FR, with brief discussions of 3D and other sensing modalities where relevant. \reviewtag{\Rthree}}

\begin{figure*}[t]
    \centering
    \includegraphics[width=1.0\linewidth]{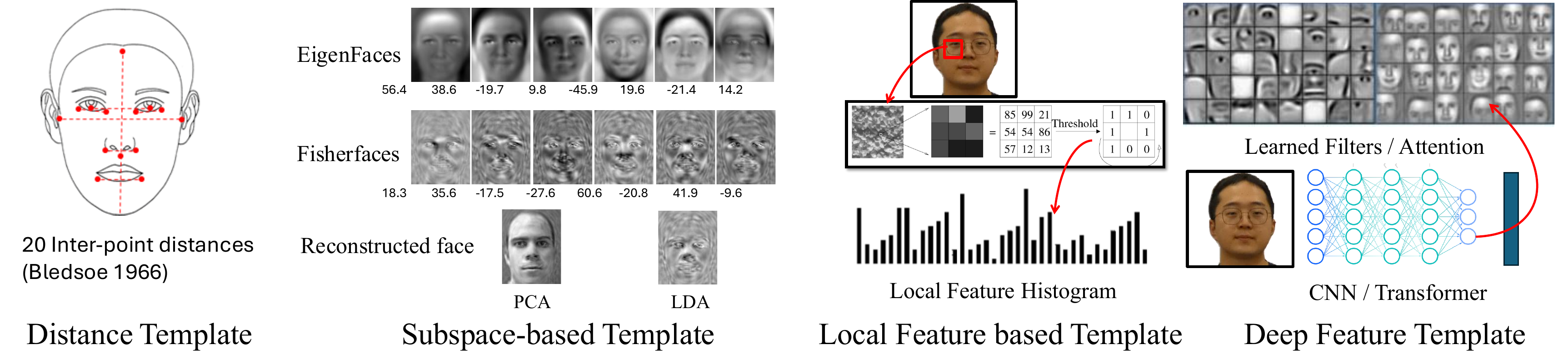}
    \caption{\textcolor{cred}{Evolution of face template representations. Early methods used geometric distances between facial landmarks (red dots and dashed lines), then subspace projections ({\it e.g.}, PCA, LDA), and local texture histograms from small patches (red square). Modern approaches employ CNNs and Transformers that learn deep feature embeddings directly from data. \reviewtag{\Rtwo}}}
    \label{fig:template_progression}
    \vspace{-4mm}
\end{figure*}

While several valuable surveys~\cite{wang2022survey,wang2021deep,guo2019survey,kortli2020face} on FR have emerged in recent years, often providing detailed catalogs of contemporary techniques or in-depth explorations of particular sub-domains, this paper offers a distinct perspective; our work spans the full 50-year evolution of the field, providing a comprehensive historical narrative that contextualizes the current State-of-the-Art (SoTA) within its broader trajectory. Recent surveys have also addressed specialized topics within the broader face analysis domain, such as 3D FR~\cite{jing20233d}, demographic bias~\cite{kotwal2025review} and face anti-spoofing~\cite{yu2022deep}, offering deep dives into these critical subfields.
In contrast, our survey maintains a strict focus on FR itself, with slight mention of important related tasks.

\section{Face Recognition Framework}

\begin{quote}
"This recognition problem is made difficult by the great variability in head rotation and tilt, lighting intensity and angle, facial expression, aging, etc."

\hfill --- Bledsoe, Chan and Bisson (1966)
\end{quote}
\label{sec:framework}

A modern FR system operates in two phases: \textit{enrollment} and \textit{recognition}\footnote{\textit{Comparison} and \textit{recognition} are used interchangeably, with recognition applying a threshold to a similarity score.}. Enrollment captures and stores facial data as a baseline, while recognition uses it to confirm or determine identity. Recognition involves two tasks: \textbf{Verification}, a one to one comparison verifying if a face matches a claimed identity; and \textbf{Identification}, a one to many search identifying a face from a database of enrolled faces.

While identification serves the broader goal of determining who someone is, it unfolds in two distinct forms: \textbf{closed retrieval} and \textbf{open search}. \textit{Closed Retrieval} assumes the probe image belongs to a known individual, ranking potential matches within a predefined set, a method long relied upon in forensic investigations and archival systems. \textit{Open Search}, on the other hand, acknowledges the unknown, forcing the system to not only rank candidates but also reject impostors when necessary. This distinction, subtle yet profound, underpins the challenge of building FR systems that are both inclusive and discerning, ensuring that recognition is not merely about finding similarities but also knowing when to say, ``this face does not belong in a dataset of interest, \textit{e.g.} a watchlist.” As size of the face databases continue to grow, FR algorithms need to be scaled for higher accuracy and speed. The largest known face database is reported to have \textcolor{cred}{50 billion facial images~\cite{clearview2025overview}} \reviewtag{\Rone, \Rthree}.

\begin{figure}
    \centering
    \includegraphics[width=1.0\linewidth]{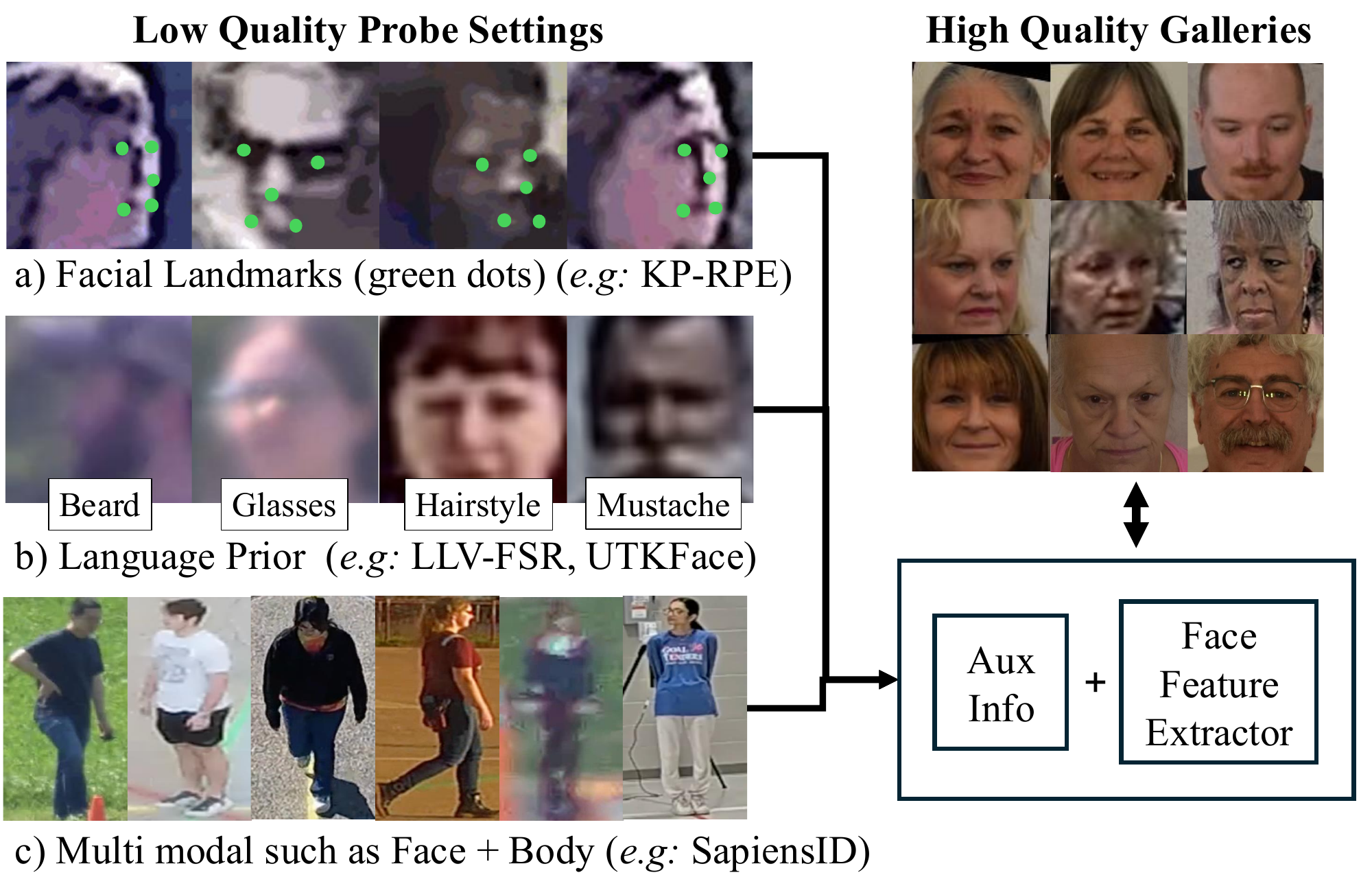}
\caption{Illustration of template enhancement by incorporating auxiliary information such as facial landmarks ({\it e.g.}, KP-RPE~\cite{kim2024keypoint}), language priors ({\it e.g.}, LLV-FSR~\cite{wang2024llv}), and multi-modal cues ({\it e.g.}, SapiensID~\cite{sapiensid}).}
    \label{fig:aux_template}
    \vspace{-4mm}
\end{figure}

The enrollment process begins with capturing a digital representation of a face (the ‘gallery’ or ‘target’ image). This raw data is then processed through a series of steps, beginning with quality assessment to ensure reliability. A crucial component is the \textit{feature extraction} stage, where salient characteristics are distilled from the facial image, creating a compact and informative ‘template’. \textcolor{cred}{This template is stored in a database and the original image can optionally be discarded for efficient comparison. \reviewtag{\Rone} } Fig.~\ref{fig:template_progression} summarizes the evolution of face templates. In some applications, the face image is also stored in addition to the template for manual adjudication. Furthermore, as depicted in Fig.~\ref{fig:aux_template}, auxiliary information such as facial landmarks, semantic attributes and multi-modal face-body cues can be integrated to enrich the template. During recognition, a feature set is extracted from the input face and compared against the stored templates using a \textit{matching function} to generate a similarity score. A decision (acceptance or rejection for verification, ranking for identification) is then made based on this score.

However, achieving robust and accurate FR is inherently challenging. The appearance of a face is remarkably variable, influenced by a multitude of factors. These \textit{intra-class or intra-person} variations encompass changes in lighting conditions, head pose, facial expression, age progression, and the presence of occlusions such as glasses, hats, or masks~\cite{lfw,calfw,agedb,cplfw}. Variations in image quality (\textit{e.g}., resolution, blur, and noise) further exacerbate the problem~\cite{ijbb,ijbc,ijbs}. Early FR systems often struggled to address these challenges, necessitating carefully controlled imaging environments with constrained pose and illumination conditions~\cite{turk1991face,ahonen2004face}.

Yet, intra-class variations cannot be examined in isolation. 
FR systems must also achieve high variance for different subjects.
This means contending with inter-class similarities, even when different individuals exhibit highly similar facial features. This includes biological cases such as identical twins and familial resemblances, where genetic similarities result in closely matching facial structures~\cite{klare2011analysis}. Moreover, non-related individuals may coincidentally look alike (so-called doppelgangers) further complicating the discrimination task~\cite{swearingen2021lookalike}. Both intra-class variation and inter-class similarity must be jointly addressed to design FR systems that are both robust and discriminative.

Modern FR systems strive to achieve invariance to intra-class variations and variance to inter-class differences~\cite{sun2014deep,schroff2015facenet,deng2019arcface,wang2017normface,wang2018cosface,huang2020curricularface,kim2022adaface}. This has been achieved through increasingly sophisticated algorithms, moving from hand-engineered features to learned representations via machine learning and, most recently, deep learning. The ability to effectively manage these sources of variation remains a central focus of ongoing research, driving the development of more resilient and reliable FR technologies. The following sections will detail the evolution of techniques used to address these challenges, from the earliest approaches to the cutting-edge methods employed today.

\begin{figure}[t]
    \centering
    \includegraphics[width=0.8\linewidth]{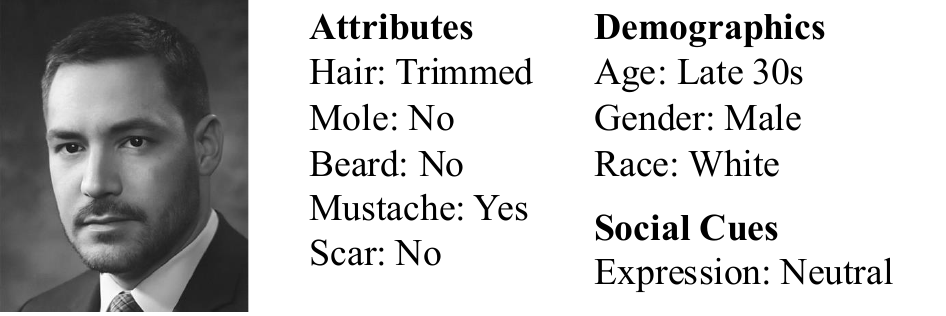}
    \caption{A breakdown of the various types of information that can be extracted from a human face. These include identity-specific features, demographic traits, soft biometrics (beard, mustache, scar), and high-level social cues such as emotion or expression.}
    \label{fig:facecontent}
    \vspace{-4mm}
\end{figure}

The human face encodes a wide spectrum of information. 
As illustrated in Fig.~\ref{fig:facecontent}, a single image can reveal identity, demographic traits, physical attributes, and social cues. 
However, deep learning-based FR systems typically do not treat these aspects independently. Instead, they amalgamate all visible cues—be it scars, expressions, or age—into a compact, high-dimensional embedding. 
While this approach has driven significant performance gains, it often comes at the cost of interpretability. The resulting features are highly discriminative but opaque, making it difficult to disentangle what specific attributes are contributing to a match decision. As FR systems become more pervasive, improving the transparency and explainability of these learned representations remains an important area of ongoing research.

\section{History of Face Recognition}
\label{sec:history}
\textbf{Early Precursors.}
The history of FR is intertwined with the broader need for reliable personal identification, initially driven by law enforcement and security concerns. The enactment of the Habitual Criminals Act in 1869 in the UK marked an early attempt to formalize the identification of repeat offenders \cite{spearman_crime_1869}. This period also saw the rise of fingerprinting, with pioneers like Henry Faulds, Francis Galton and Edward Henry recognizing the uniqueness and potential of minutiae points for individual identification \cite{faulds1880skin}. 
\textcolor{cred}{These pre-digital efforts established the conceptual foundation for biometric identification, which would later evolve into automated computer vision systems. \reviewtag{\Rthree}}

\textbf{The First Automated Systems (1960s–1980s).}
Early computer-based FR systems emerged in the 1960s, notably with the work of Bledsoe, who used manually annotated facial landmarks and feature distances to identify individuals \cite{bledsoe1966man}. 
\textcolor{cred}{Kelly ~\cite{kelly1970visual} attempted to automate facial identification by using computer vision to measure distances between key facial landmarks, such as the eyes and nose. \reviewtag{\Rthree}}
A major milestone followed with Kanade’s 1973 dissertation, which presented the first fully automated FR system \cite{kanade1974picture}. 

\textbf{The Rise of Feature Engineering (1990s–2000s).}
The 1990s mark a paradigm shift in FR, moving from handcrafted geometric features to holistic, appearance-based representations. 
A seminal contribution is the Eigenfaces by Turk and Pentland~\cite{turk1991face}, which leverages PCA to represent faces as linear combinations of orthogonal basis images. This approach enables more compact and discriminative facial representations. 
However, Eigenfaces exhibit limitations in handling variations in lighting and facial expression. To address this, Belhumeur {\it et al.}~\cite{belhumeur1997eigenfaces} introduce Fisherfaces, which apply linear discriminant analysis (LDA) to better separate individuals, improving robustness under varying illumination. This PCA-versus-LDA debate is further explored by Martinez and Kak~\cite{martinez2001pca}, who highlight the strengths and weaknesses of both in practical scenarios.

\textcolor{cred}{Building on holistic approaches, researchers aimed to model both facial appearance and shape variability. Lanitis, Taylor, and Cootes (1995)~\cite{lanitis1995unified} proposed a PCA-based framework combining geometry and grey-level appearance for automatic face analysis. Moghaddam and Pentland (1997)~\cite{moghaddam1997probabilistic} introduced a probabilistic eigenspace, modeling object classes as Gaussian or Mixture of Gaussians and framing recognition as maximum likelihood estimation. These works established a unified, data-driven foundation for appearance-based recognition. \reviewtag{\Rthree}
}

Model-based techniques such as Elastic Bunch Graph Matching~\cite{wiskott2022face} provide pose-invariant recognition by encoding facial landmarks through a graph-based structure, bridging the gap between rigid appearance models and deformable representations. Complementing these efforts are texture-based descriptors such as Local Binary Patterns (LBP)~\cite{ahonen2004face} and Scale-Invariant Feature Transform (SIFT)~\cite{lowe2004distinctive}, which mitigate sensitivity to lighting and expression by capturing local structural patterns.

A critical breakthrough in face detection in images emerges with the Viola–Jones algorithm~\cite{viola2001rapid}, enabling real-time detection by Haar-like features and boosting. 
This work opens doors for practical applications in surveillance and consumer electronics. 
Around the same period, Heisele {\it et al.}~\cite{heisele2007component} proposed a component-based framework, integrating part-based local features to enhance robustness against occlusion and pose variation, thereby reinforcing the shift toward modular and discriminative feature engineering.

\textcolor{cred}{The FERET program~\cite{phillips1998feret} advanced evaluation protocols by introducing standardized datasets and methods for comparing FR algorithms. Its gallery probe design and reporting metrics became key benchmarks in the 1990s and early 2000s, forming the foundation for empirical progress. \reviewtag{\Rfour}
}

\textbf{Deep Learning Paradigm (2010s–Present).} 
The advent of deep learning~\cite{krizhevsky2012imagenet,he2016deep} in the 2010s revolutionized the field. This paradigm shift is fueled by innovations in neural network architectures and the availability of large-scale datasets for training and evaluating the networks. 
Landmark papers like AlexNet~\cite{krizhevsky2012imagenet} and ResNet~\cite{he2016deep} demonstrate the power of convolutional networks for image recognition, paving the way for their adoption in FR. The ImageNet dataset~\cite{krizhevsky2012imagenet} provides a crucial resource for pre-training these large models, which were then fine-tuned for FR tasks.

Early pioneering works like DeepFace \cite{taigman2014deepface} and FaceNet \cite{schroff2015facenet} demonstrate the potential of deep learning for FR, achieving near-human performance on benchmark datasets. DeepFace utilizes a large-scale dataset (4M images and 4K subjects) of facial images to train a deep neural network for face verification, while FaceNet introduces a unified embedding space for FR and face grouping.

A key area of innovation in deep FR has been the development of loss functions (Sec.~\ref{sec:loss}). These loss functions are designed to improve the discriminative power of the learned features, enabling more accurate FR. Notable examples include NormFace \cite{wang2017normface}, SphereFace \cite{liu2017sphereface}, CosFace \cite{wang2018cosface}, ArcFace \cite{deng2019arcface}, CurricularFace \cite{huang2020curricularface} and Adaface \cite{kim2022adaface}, each introducing novel approaches to margin-based learning.

The performance of deep FR models is also highly dependent on the availability of large-scale training datasets (Sec.~\ref{sec:dataset}). Several large-scale face datasets have been developed to train and evaluate FR models, including CASIA-WebFace~\cite{casia}, VGGFace~\cite{vgg}, MS1M~\cite{vgg}, and WebFace260M~\cite{webface}. These datasets provide a diverse range of facial images with varying pose, illumination, expression and age, enabling the training of robust FR models.

\begin{figure}[t]
    \centering
    \includegraphics[width=1.0\linewidth]{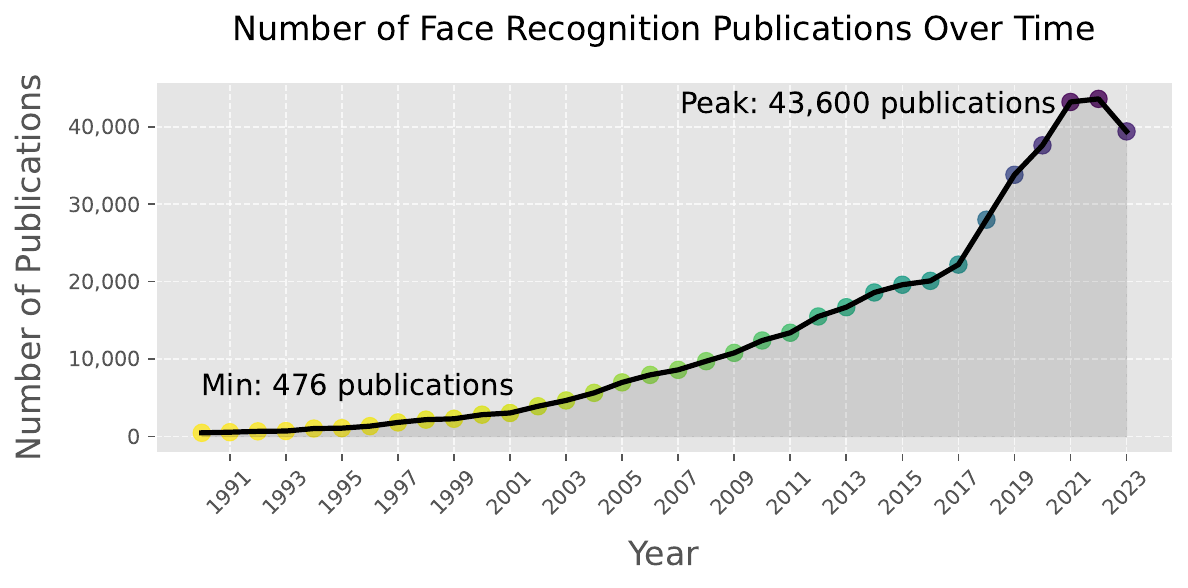}
    \vspace{-6mm}
    \caption{Number of FR publications over time. Research activity in the field grew steadily until the early 2010s, followed by an explosive increase coinciding with the rise of deep learning. The peak in 2022 reflects the technology’s mainstream adoption, though recent years suggest a slight cooling-off period in publications. However, in terms of deployments, FR continues to gain momentum. The global FR market size was valued at USD 7.73 billion in 2024. The market is projected to grow from USD 8.83 billion in 2025 to USD 24.28 billion by 2032~\cite{fortune2025facial}.}
    \label{fig:fr_publication_trend}
    \vspace{-4mm}
\end{figure}

With the advent of deep learning, advanced architectures (Sec.~\ref{sec:arch}) with billions of parameters that are trained on increasingly large face datasets bring about unprecedented improvements in accuracy and robustness. Today, FR stands as one of the most successful applications of deep neural networks in computer vision and AI, as shown in Fig.~\ref{fig:fr_publication_trend}.

\section{Advances in Deep Face Recognition}
\label{sec:advances}
Deep face recognition has advanced through improved loss functions, large diverse datasets, and better neural architectures, enabling more discriminative representations.

\subsection{Loss Functions}
\label{sec:loss}

\textcolor{cred}{
Before deep learning, FR relied on analytical methods lacking desired discriminative power. Eigenfaces~\cite{turk1991face} used Principal Component Analysis to maximize variance, and Fisherfaces~\cite{belhumeur1997eigenfaces} applied Linear Discriminant Analysis to enhance class separability, both requiring costly covariance matrix diagonalization~\cite{martinez2001pca}. Deep learning replaced these with gradient based optimization and specialized loss functions~\cite{wang2017normface,liu2017sphereface,wang2018cosface,deng2019arcface,huang2020curricularface,kim2022adaface} that directly enforce inter-class separability and intra-class compactness. \reviewtag{\Rthree}
}

The choice of loss function is critical in training deep FR models, as it directly guides the network to learn discriminative feature embeddings suitable for distinguishing between a vast number of identities. Among the key innovations that drove the advancement in FR, the highest number of publications have come from the advances in the loss function. Several distinct paradigms for loss function design have emerged.

\textbf{Contrastive Learning Approaches:} One major family of loss functions employs contrastive learning principles, directly shaping the embedding space by optimizing relative distances between samples. The seminal FaceNet~\cite{schroff2015facenet} introduces the triplet loss, designed to ensure that an anchor sample's embedding is closer to its positive (same identity) counterpart than to any negative (different identity) sample by a predefined margin, typically in the Euclidean space.

While effective, triplet loss faces challenges in sampling informative triplets. Many randomly chosen triplets provide weak gradient signal, making training inefficient or necessitating complex hard-negative mining strategies. To address this, proxy-based methods are proposed. Techniques like Proxy Anchor Loss~\cite{kim2020proxy}, originating from general deep metric learning, associate each class with learnable proxies (representative vectors), simplifying the loss computation by comparing samples to these proxies rather than exhaustively searching for pairs or triplets within a batch.

\begin{figure}
    \centering
    \includegraphics[width=0.85\linewidth]{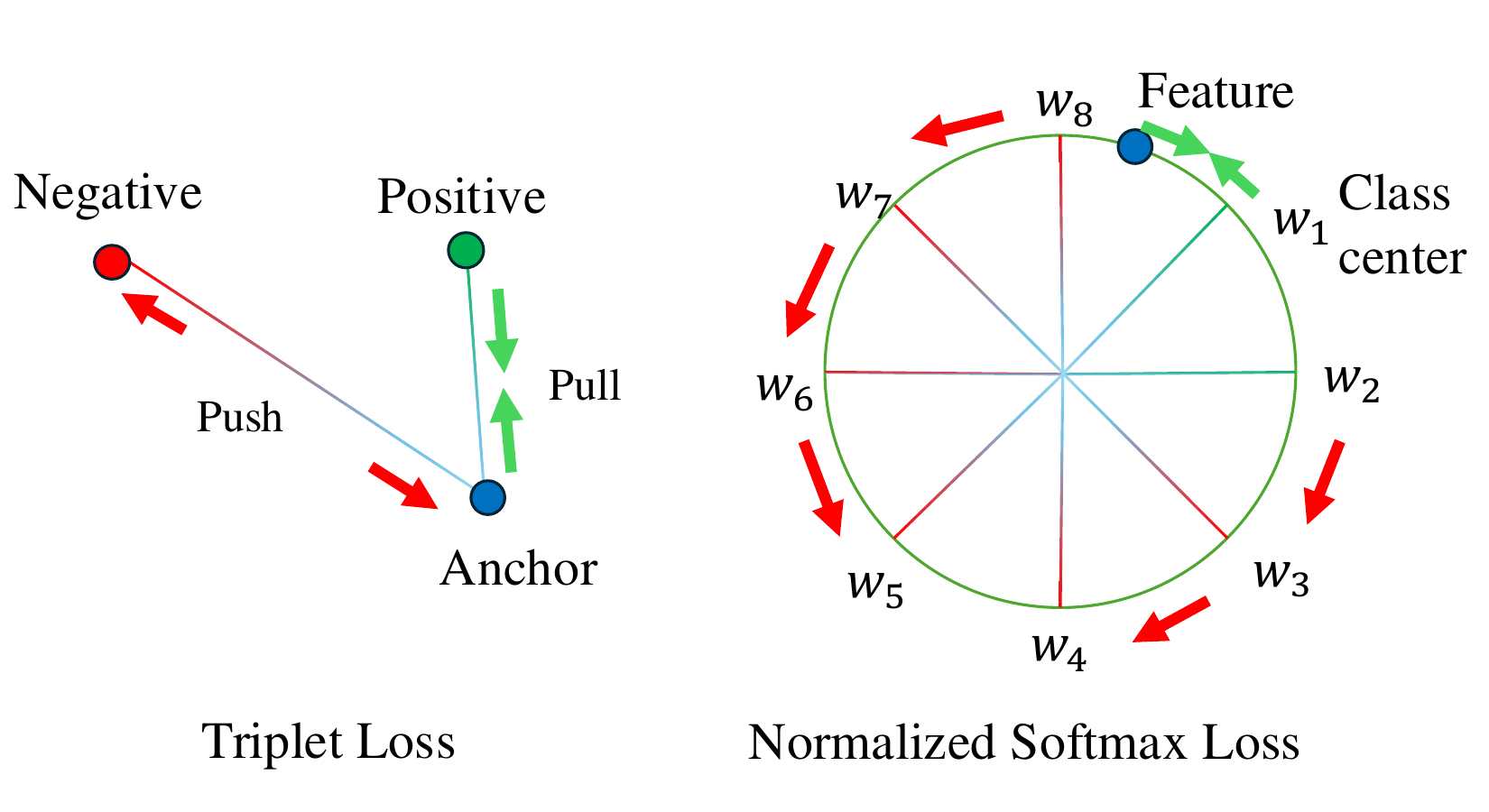}
    \caption{Comparison of loss function paradigms in deep FR. Left: Triplet loss in contrastive learning reduces intra-class distance (pull positive training samples closer to anchor) while increasing inter-class distance (push negative training samples away). Right: Normalized Softmax loss maps features and class weights onto a hypersphere, optimizing angular distances to enhance inter-class separability and intra-class compactness.}
    \label{fig:loss_comparison}
    \vspace{-4mm}
\end{figure}

\begin{table*}[t]
    \centering
    \caption{Summary of deep FR methods focusing on their loss functions, with their key advantages and limitations.}
    \resizebox{0.9\linewidth}{!}{
        \scriptsize
            \renewcommand{\arraystretch}{1.2} 

    \begin{tabular}{|l|c|l|l|}
        \hline
        \textbf{Name} & \textbf{Year} & \textbf{Pros} & \textbf{Cons} \\
        \hline
        DeepID2+~\cite{sun2015deeply} & 2014 & Joint ID + verification loss for robust features & Complex training with dual supervision losses \\
        \hline
        CenterFace~\cite{centerloss} & 2016 & Center loss enhances intra-class compactness & No explicit inter-class separation; needs tuning \\
        \hline
        SphereFace~\cite{liu2017sphereface} & 2017 & Angular margin enforces hyperspherical separation & Training instability from angular multiplicity \\
        \hline
        L2-Face~\cite{ranjan2017l2} & 2017 & L2 norm constraint improves angular discrimination & Needs careful radius tuning; no margin enforcement \\
        \hline
        ArcFace~\cite{deng2019arcface} & 2018 & Additive angular margin boosts inter-class separation & Fixed margin may hurt low-quality samples \\
        \hline
        CosFace~\cite{wang2018cosface} & 2018 & Cosine margin improves class separability stably & Uniform margin not adaptive; needs tuning \\
        \hline
        SeqFace~\cite{hu2018seqface} & 2018 & Sequence-aware loss improves temporal supervision & Needs sequence data; dual loss increases complexity \\
        \hline
        Git Loss~\cite{calefati2018git} & 2018 & Unified softmax + center loss boosts discrimination & Extra tuning and complexity with marginal gain \\
        \hline
        MagFace~\cite{meng2021magface} & 2021 & Feature norm models quality for adaptive margin & Complex loss and quality-norm assumptions \\
        \hline
        AdaFace~\cite{kim2022adaface} & 2022 & Dynamic margin based on feature norm quality & Relies on norm-quality link and tuning \\
        \hline
        ElasticFace~\cite{boutros2022elasticface} & 2022 & Elastic margin adapts to feature variability & Stochastic margins add tuning and training cost \\
        \hline
        UniFace~\cite{zhou2023uniface} & 2023 & Similarity threshold improves verification alignment & Global constraints increase optimization cost \\
        \hline
        UniTSFace~\cite{jia2023unitsface} & 2023 & Sample-to-sample loss optimizes verification & Pairwise loss and threshold learning cause overhead \\
        \hline
        UCFace~\cite{ahn2024uncertainty} & 2024 & Uncertainty and probability density aware contrastive learning & Cannot be used by itself, must be accompanied by margin loss \\
        \hline
        LAFS~\cite{sun2024lafs} & 2024 & Landmark based SSL pretraining helps FR & Loss depends on pretrained model and the landmark quality. \\\hline
    \end{tabular}
    }
    \label{tab:marginlosssummary}
    \vspace{-2mm}
\end{table*}

Further refining the contrastive approach, Supervised Contrastive Learning (SupCon)~\cite{khosla2020supervised} generalizes the loss to leverage all positive samples available for an anchor within a batch, contrasting them against all negative samples. This more data-efficient approach has been successfully applied to FR, for instance, in UCFace~\cite{ahn2024uncertainty}. Other works adapt contrastive ideas for specific goals: Open-Set Biometrics~\cite{su2024open} focuses on improving open-set performance by explicitly minimizing scores between non-mated pairs, while CAFace~\cite{kim2022cluster} uses a contrastive-style cosine similarity loss to enforce consistency between embeddings of low-quality images and their high-quality counterparts, promoting quality invariance. Related efforts also explore optimizing embedding spaces to better align with recognition objectives, such as in~\cite{tu2009optimizing}, where features are learned to directly improve performance metrics for identification and verification separately.

\textbf{Margin-based Softmax Losses:}
A dominant and highly successful approach in deep FR involves modifying the standard softmax cross-entropy loss to directly enhance feature discriminability. The core motivation is to learn embeddings that exhibit smaller intra-class variations (same person are close together) while simultaneously maximizing inter-class separation (different people are far apart).

The standard softmax loss, often used as a baseline in classification tasks, is formulated for a sample $\mathbf{x}_i$ with feature embedding $\mathbf{z}_i \in \mathbb{R}^d$ belonging to the $y_i$-th class as:
\begin{equation}
\mathcal{L}_{CE}(\mathbf{x}_i)
= -\log\tfrac{\exp(\mathbf{W}_{y_i}^\top\mathbf{z}_i+b_{y_i})}
{\sum_{j=1}^C\exp(\mathbf{W}_j^\top\mathbf{z}_j+b_j)},
\label{eq:softmax}
\end{equation}
where $\mathbf{W}_j$ is the weight vector for the $j$-th class, $b_j$ is the bias term, and $C$ is the total number of classes or identities in the training set. While effective for classification, this formulation doesn't explicitly enforce the metric learning objective crucial for FR where we encounter identities not seen during training.

An early work moving in this direction is Center Loss~\cite{centerloss}, which adds an auxiliary loss term to the standard softmax. This term penalizes the Euclidean distances between the deep features and their corresponding learned class centers, directly encouraging intra-class compactness. A significant breakthrough comes with the normalization of both feature embeddings ($\|\mathbf{z}_i\|=1$) and classification weights ($\|\mathbf{W}_j\|=1$, and setting $b_j=0$). This reformulation, pioneered by SphereFace~\cite{liu2017sphereface}, maps the optimization problem onto a hypersphere where the dot product $\mathbf{W}_j^T \mathbf{z}_i$ becomes equivalent to $\cos\theta_j$, the cosine of the angle between the feature vector $\mathbf{z}_i$ and the weight vector $\mathbf{W}_j$. A scaling factor $s$ is typically introduced to control the radius of the hyperspherical feature space. The loss then becomes:
\begin{equation}
\mathcal{L}_{cos}(\mathbf{x}_i)
= -\log\tfrac{\exp(s\!\cdot\!\cos\theta_{y_i})}
{\sum_{j=1}^C \exp(s\!\cdot\!\cos\theta_j)}.
\label{eq:cos_softmax}
\end{equation}

Building on this normalized angular space, the key innovation is the introduction of explicit margins to make the learning objective more stringent. CosFace~\cite{wang2018cosface} introduces an additive cosine margin ($m$) by modifying the target logit to $s \cdot (\cos\theta_{y_i} - m)$. ArcFace~\cite{deng2019arcface} proposes an additive angular margin ($m$) by modifying the target angle itself, resulting in a target logit of $s \cdot \cos(\theta_{y_i} + m)$. Both approaches effectively create a decision boundary gap, forcing learned features for the same identity to cluster more tightly in the angular space, thereby significantly improving discriminative power. A visual comparison of contrastive triplet loss and margin-based normalized softmax loss is illustrated in Fig.~\ref{fig:loss_comparison}, highlighting how each paradigm optimizes the embedding space to enhance FR performance.

Subsequent research further refines these margin-based concepts. MagFace~\cite{meng2021magface} proposes leveraging the magnitude of the feature vector (before normalization) as an indicator of face image quality, incorporating an auxiliary loss to promote larger magnitudes for higher-quality samples. AdaFace~\cite{kim2022adaface} addresses the challenge posed by low-quality or difficult samples by introducing an adaptive margin function. It dynamically adjusts the margin stringency based on image quality indicators, reducing the negative impact of potentially unrecognizable faces in the training process.

These advancements in margin-based softmax losses lead to remarkable performance gains, pushing verification accuracy on high-quality benchmarks like LFW~\cite{lfw} and CFP-FP~\cite{cfpfp} towards saturation (often exceeding 99\%). This success shifts the community's focus towards improving performance in more challenging, unconstrained scenarios, particularly those involving low-quality images, as represented by benchmarks like IJB-S,TinyFace or BRIAR~\cite{ijbs,cheng2018low,BRIAR}. \textcolor{cred}{The IJB-S and BRIAR datasets are limited-access evaluation sets developed under U.S. Government research programs and are distributed selectively for research use. \reviewtag{\Rtwo}}

Further refinements continued to explore margin dynamics; for example, ElasticFace~\cite{boutros2022elasticface} introduces randomized margins for greater flexibility during training, while UniFace~\cite{zhou2023uniface} proposes the Unified Cross-Entropy (UCE) loss specifically aiming to guarantee a clear separation threshold between positive and negative pairs.

Margin-based softmax variants~\cite{deng2019arcface,kim2022adaface,zhou2023uniface} currently dominate SoTA results. Contrastive methods remain a helpful auxiliary loss, on top of margin-based softmax losses. A summary of various loss functions are shown in Tab.~\ref{tab:marginlosssummary}.

\textbf{Auxiliary Losses for Interpretability and Distillation:}
Beyond optimizing the core embedding space based purely on identity labels, another category of loss functions incorporates auxiliary objectives to achieve specific goals, such as enhancing model interpretability or improving performance in challenging conditions like low resolution. These losses often supplement the primary identity discrimination loss.

A significant effort has focused on improving model interpretability, {\it e.g.}, understand \textit{how} the network makes decisions. Towards this, Yin {\it et al.}~\cite{yin2019towards} propose spatial and feature activation diversity losses. These encourage the network to learn more structured representations where different spatial activations may correspond to different facial aspects, while also making these interpretable features discriminative and robust to occlusions. Similarly, the Explainable Channel Loss (ECLoss), also framed as Activation Template Matching Loss~\cite{Lin2024ECLoss}, encourages specific channels within convolutional layers to specialize in detecting distinct facial parts ({\it e.g.}, eyes) without explicit part annotations, thereby providing a direct interpretation of channel function.

Knowledge distillation (KD) offers another avenue, often targeting specific challenges. For low-resolution FR, Attention Similarity KD (A-SKD)~\cite{shin2022teaching} transfers teacher attention maps to guide the student's focus. For efficient large-scale training, Li {\it et al.}~\cite{li2023rethinking} develop feature-based KD techniques, like reverse distillation, that importantly remove the need for identity supervision for the student, saving memory while addressing the teacher-student 'intrinsic gap'. These examples highlight the use of specialized loss functions and KD strategies to imbue models with desirable properties like explainability, robustness, or training efficiency, complementing the primary task of recognition.

\begin{figure}[t]
\centering
\begin{minipage}[t]{1.0\linewidth}
    \centering
    \includegraphics[width=\linewidth]{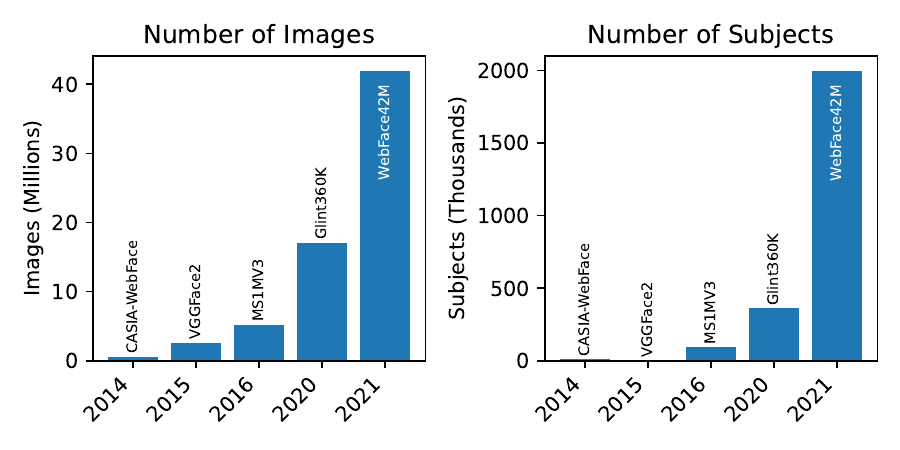}
    \vspace{-6mm}
    \caption{Plots showing the growth of FR datasets over time. The left illustrates the number of images (in millions), and the right shows the number of subjects (in thousands) for each dataset.}
    \label{fig:datasetinfo}
    \vspace{-1mm}
\end{minipage}

\hfill
\vspace{2mm}
\begin{minipage}[t]{1.0\linewidth}
    \centering
    \includegraphics[width=\linewidth]{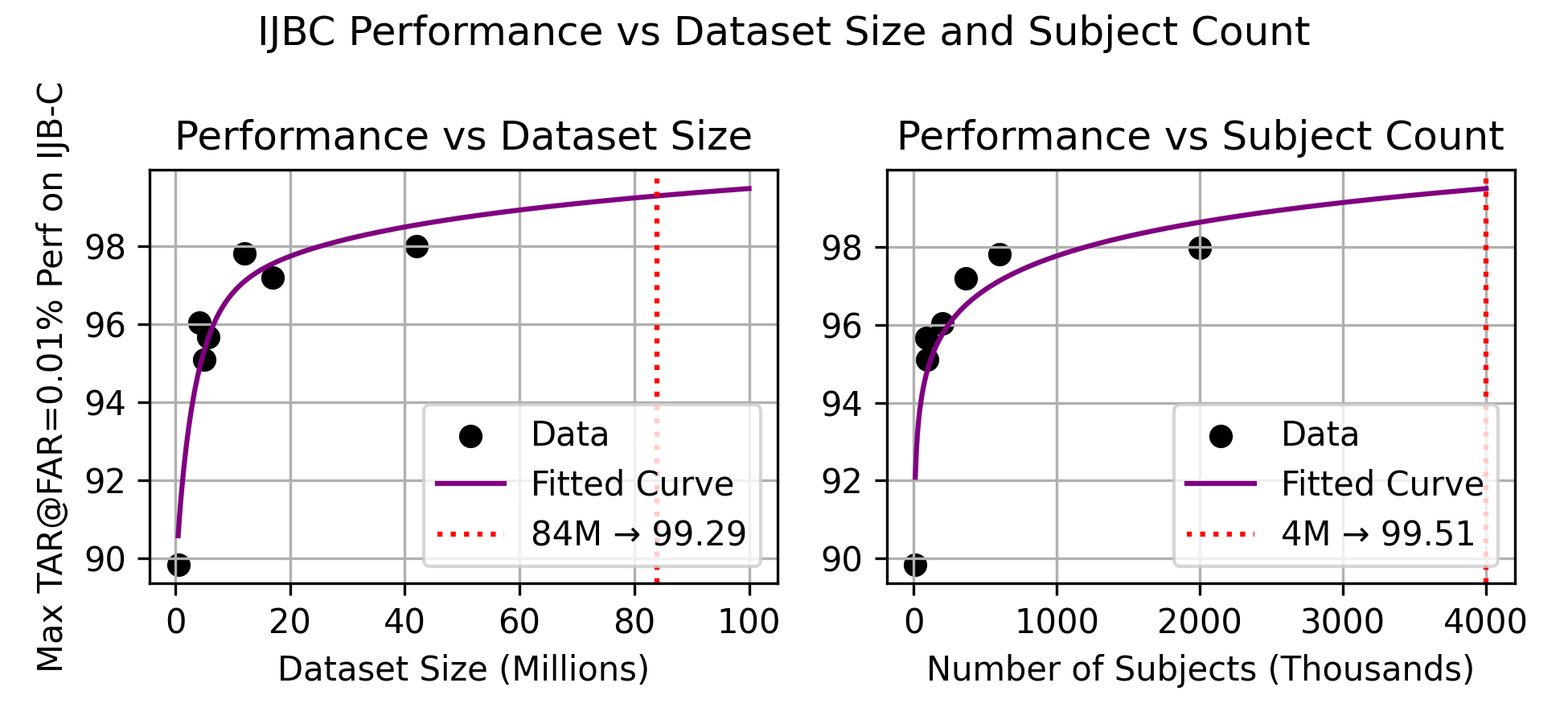}
    \vspace{-6mm}
    \caption{Recognition performance on IJB-C dataset as a function of training dataset size (left) and training number of subjects (right). The dots show the the best publically available algorithms' performance for the given training dataset. Curves are fitted using the logarithmic function. Both increasing the number of images and expanding subject diversity significantly improve performance. However, the performance begins to saturate around 42M images and 2M subjects, suggesting diminishing returns at larger scales. While further gains are still possible, it may require novel embeddings.}
    \label{fig:ijbc}
\end{minipage}
\vspace{-4mm}
\end{figure}

\subsection{Datasets}

\textcolor{cred}{
Before deep learning, FR was limited by small datasets. Prior to the 1996 FERET database, which had 1,199 individuals and became a major benchmark~\cite{phillips1998feret,phillips2000feret}, most studies used fewer than 50 subjects, restricting generalization under unconstrained conditions. These small datasets sufficed for classical approaches like PCA based Eigenfaces~\cite{turk1991face}, which relied on analytical solutions. The advent of deep learning~\cite{krizhevsky2012imagenet,taigman2014deepface,schroff2015facenet} created a vast need for data, as neural networks require far more samples to learn model parameters~\cite{vgg,ms1m,webface}. \reviewtag{\Rthree,\Rfour}
}

\label{sec:dataset}
The availability and scale of training data have been pivotal factors driving the remarkable progress in deep learning~\cite{krizhevsky2012imagenet}. Publicly available large-scale face datasets ({\it e.g.}, MS-Celeb-1M, VGGFace2) spurred rapid advancements in the mid-2010s. Fig.~\ref{fig:datasetinfo} provides an overview of several influential datasets commonly used in the field, detailing the number of images and unique identities they contain. A clear trend emerges from this summary: a dramatic increase in dataset size over time. Early benchmark datasets like CASIA-WebFace~\cite{casia} offer around half a million images from ten thousand subjects. In contrast, subsequent collections such as MS1MV2/V3~\cite{ms1m}, Glint360K~\cite{an2021partial}, and particularly the WebFace series~\cite{webface}, have pushed these numbers significantly higher, culminating in WebFace42M with over 42 million face images spanning 2 million identities. This growth reflects the community's understanding that larger and more diverse datasets are crucial for training accurate and generalizable FR models. \textcolor{cred}{It is important to note that MS-Celeb-1M and VGGFace2 have been discontinued by the dataset creators.}\reviewtag{\Rone} Some popular public datasets and pretrained model checkpoints are available in this
\href{https://github.com/deepinsight/insightface}{link}.

In Fig.~\ref{fig:ijbc}, we show the FR performance on IJB-C at TAR@FAR=0.01\% with varied dataset size and number of subjects. The performance is taken as the maximum of FR algorithms that were trained on the particular dataset. And we fit a curve to see the trend. We observe that both increasing dataset size and subject number lead to substantial improvements in performance. However, the trend indicates a saturation point around 42M images or 2M subjects, beyond which additional data yields diminishing returns.

It is important to note the origin and labeling methodology of many of these large-scale datasets. A significant portion, including prominent datasets like MS-Celeb-1M and the WebFace series, are curated by collecting images from publicly accessible sources on the Internet, often leveraging search engines or social media platforms. Consequently, the identity labels associated with these images are frequently ``pseudo-labels," because web searches of celebrities may return different subject images. Due to the volumn of these datasets, the labels are generated through automated clustering algorithms or matching, rather than manual verification. While efforts are made to clean and refine these labels, noise and inaccuracies can persist. Some approaches, like that used for WebFace260M~\cite{webface}, employ iterative self-labeling and retraining of specialized labeler models to improve the quality of these pseudo-labels over multiple cycles. Since benchmark datasets~\cite{lfw,cfpfp,agedb,ijbb,ijbc} are also curated from public web, training datsets need to ensure that the identities in training and test sets do not overlap. 

Also, the practice of web-scraping facial images raises significant privacy and ethical concerns within the research community and society at large~\cite{wang2024beyond}, as individuals may not have provided explicit consent for their images to be used in this manner for developing and training recognition systems. This issue remains an active area of discussion and necessitates careful consideration of data governance and ethical guidelines moving forward.

In addition to the large-scale 2D image datasets collected primarily from the web, the field also utilizes 3D face datasets~\cite{blanz1999morphable,paysan2009bfm,gerig2018morphable,booth2018large,ploumpis2019combining,abrevaya2018multilinear,albrecht2013statistical,li2017learning,ranjan2018generating,li2020learning,wang2022faceverse,schwartz2010inface}. These datasets capture the geometric structure of the face, often along with texture information, using specialized acquisition techniques like 3D scanners, structured light, or multi-camera stereo systems. A key distinction is that 3D datasets are typically collected under controlled laboratory settings with the explicit consent of the participants. This controlled acquisition allows for high-quality, precise capture of facial shape in RGBd (depth), which can offer inherent robustness advantages against variations in pose and illumination compared to 2D images. 

3D face recognition has also emerged as a parallel research track, with dedicated benchmarks and evaluation campaigns~\cite{phillips2005overview}. Notably, datasets such as FRGC~\cite{phillips2005overview} and Lock3DFace~\cite{zhang2016lock3dface}, which leverages low-cost RGB-D sensors like Kinect, have been widely used in the community to advance recognition algorithms under realistic conditions.

However, the process of 3D data acquisition is significantly more complex, time-consuming, and expensive. Consequently, the volume of available 3D face data, both in terms of the number of scans and the number of unique subjects, is substantially smaller compared to the massive scale achieved by web-scraped 2D datasets. This difference in scale limits the use of 3D face datasets for training the deep models that dominate current FR research, although they remain valuable for specific research tasks, evaluation, and applications where 3D information is critical. 
For general FR applications, such as in law enforcement, immigration, or airport screening, RGB cameras offer a more practical solution considering legacy databases and return on investment.

\subsection{Neural Network Architectures}
\label{sec:arch}
\noindent\textbf{CNN Architectures in FR: }
The backbone neural network architecture plays a crucial role in extracting discriminative features from face images. 
The revolution brought by deep learning in computer vision, largely initiated by AlexNet~\cite{krizhevsky2012imagenet} on the ImageNet challenge, quickly permeates the field of FR. 
Early deep FR models adapt existing CNN architectures designed for general object recognition.

Architectures like GoogLeNet~\cite{szegedy2015going} demonstrate the power of increased network depth and led to its adoption in FaceNet~\cite{schroff2015facenet}. 
The introduction of Residual Networks (ResNets)~\cite{he2016deep} which addresses the vanishing gradient problem in very deep networks through the use of residual connections (shortcuts) leads to training of much deeper models ({\it e.g.}, ResNet-50, ResNet101, ResNet-152).
Variants of ResNet ({\it e.g.}, SE blocks~\cite{hu2018squeeze}), become the popular backbone for many SoTA FR systems developed in the late 2010s~\cite{deng2019arcface, wang2018cosface}. 
ArcFace~\cite{deng2019arcface}'s adoption of input size $112 \times 112$ leads to the widely used IR-ResNet backbones which removed first downsampling blocks to compensate for the small resolution. Fig.~\ref{fig:facesize} shows the progression of facial image sizes in the FR datasets.

Facial alignment is a crucial preprocessing step in FR systems, ensuring that key facial features (\textit{i.e.}~eyes, nose, and mouth) are consistently positioned across different images. 
Earlier FR datasets~\cite{casia} utilize Multi-task Cascaded Convolutional Network (MTCNN)\cite{zhang2016joint}, which jointly performs face detection and landmark localization via a series of cascaded networks. 
With the advent of single-stage detectors like SSD~\cite{liu2016ssd}, more efficient and accurate methods emerge. 
Notably, RetinaFace~\cite{deng2019retinaface} has become a popular solution, offering precise face detection and alignment. 
When trained on strong datasets such as WiderFace~\cite{yang2016wider} and paired with an improved backbone, RetinaFace is a robust choice for preprocessing large-scale face datasets~\cite{webface}. 
Some example alignments are shown in the last row of Fig.~\ref{fig:facesize}.

\begin{figure}[t]
    \centering
    \includegraphics[width=0.9\linewidth]{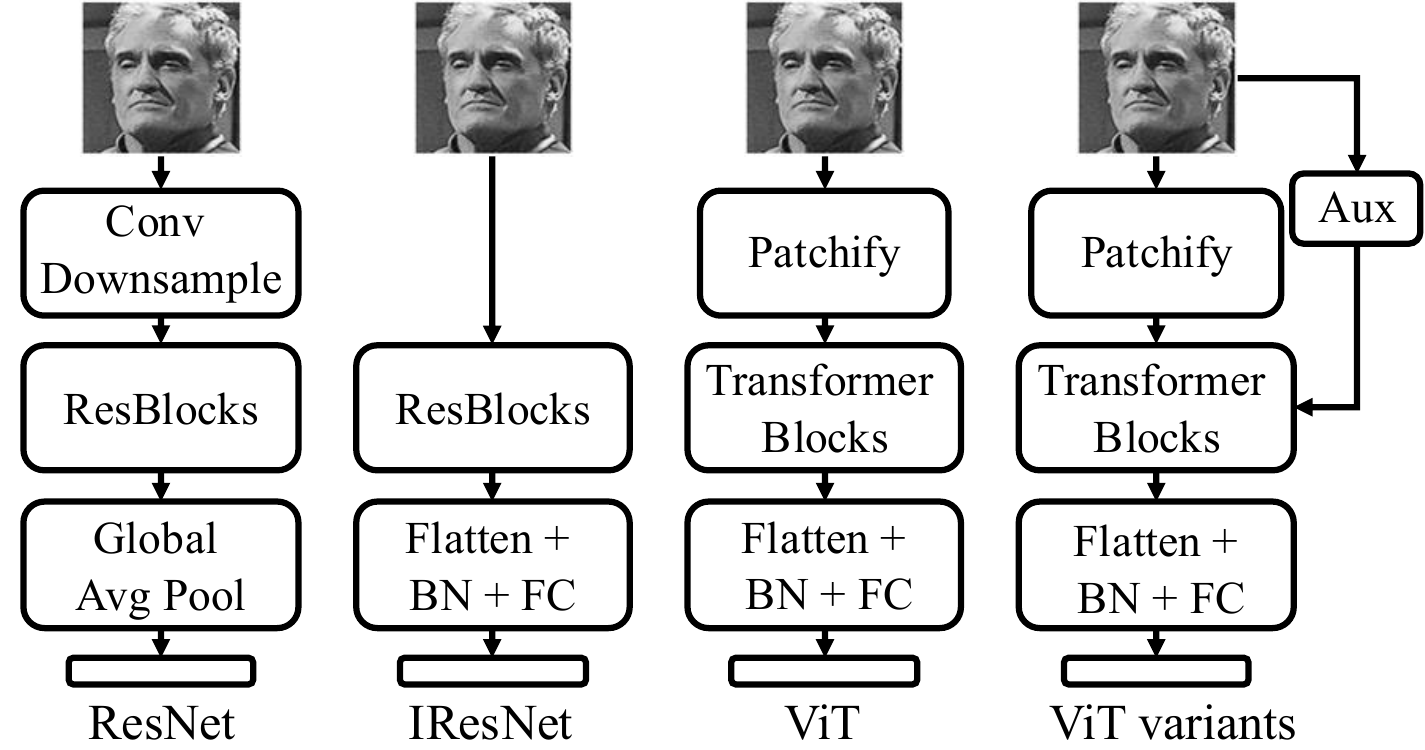}
    \caption{Comparison of architectures used in FR. From left to right: ResNet~\cite{he2016deep} is an architecture used for classification~\cite{krizhevsky2012imagenet}. IResNet~\cite{deng2019arcface} modifies this by removing downsampling and using feature flattening, batch normalization (BN), and fully connected (FC) layers that are helpful for metric learning~\cite{luo2019bag}. Vision Transformer~\cite{dosovitskiy2020image} (ViT) replaces convolutions with a patchify operation and transformer blocks; ViT variants further extend this by incorporating auxiliary information such as facial keypoints~\cite{deng2019retinaface,kim2024keypoint} to improve learning.}
    \label{fig:arch}
    \vspace{-4mm}
\end{figure}

\noindent\textbf{Vision Transformers in FR: } 
Mirroring trends in natural language processing and broader computer vision, Vision Transformers (ViTs)~\cite{dosovitskiy2020image} have emerged as a powerful alternative to CNNs. ViTs process images by dividing them into patches, and feeding the resulting sequence into a Transformer encoder~\cite{vaswani2017attention}. The self-attention mechanism within Transformers allows the model to weigh the importance of different image patches globally, potentially capturing long-range dependencies that might be missed by the local receptive fields of CNNs. ViTs have also shown great performance in FR domains~\cite{rodrigo2024comprehensive,qin2023swinface,kim2024keypoint}. And adoption of ViT in FR implies adoption of advances around ViT. SwinFace~\cite{qin2023swinface} is an application of Swin Transformer~\cite{liu2021swin}. KP-RPE\cite{kim2024keypoint} integrates facial landmarks into relative position encodings in ViT, improving robustness to pose variations.

Empirically, compared to ResNets, training ViT on FR models entails more augmentations and requires larger-scale training set~\cite{kim2024keypoint}. 
Fig.~\ref{fig:arch} shows how FR models have changed over time, from ResNets to newer ViT models that use attention and extra information to recognize faces better.

While both CNNs, particularly ResNet variants like IR-SE models, and Vision Transformers (ViTs) have demonstrated SoTA performance, there is no single definitively best architecture for all FR tasks. ViTs have shown potential for marginally higher accuracy on some benchmarks, especially when trained on extremely large datasets, leveraging their capability to capture global image dependencies. However, they often necessitate more extensive training data and sophisticated augmentation strategies. ResNets remain highly competitive, often providing a more efficient balance between performance and training/inference cost, particularly with moderate-sized datasets. The optimal choice often depends on the specific application's constraints, including dataset size, compute resources, and deployment setting.

FR model deployment depends heavily on computational demands, mainly measured by FLOPs and model size. For example, IResNet50 needs 12.62 GFLOPs and has 43.59M parameters, while IResNet101 uses 24.19 GFLOPs and 65.15M parameters. ViT models are more demanding—ViT Small has 17.42 GFLOPs and 95.95M parameters, and ViT Base requires 24.83 GFLOPs with 114.87M parameters. On consumer GPUs such as Nvidia 3090, IResNet50 can process over 1400 images/second, while ViT Base handles about 640 images/second. Unlike academic research, industry models or government vendor models~\cite{liu2024farsight} can use model ensembles, further increasing the load. Preprocessing steps like face detection and alignment add to the computational cost.
Note that ViT backbones can benefit from research that speed up ViT inference~\cite{dao2022flashattention,xFormers2022}.

\begin{figure}[t]
    \centering
    \includegraphics[width=0.8\linewidth]{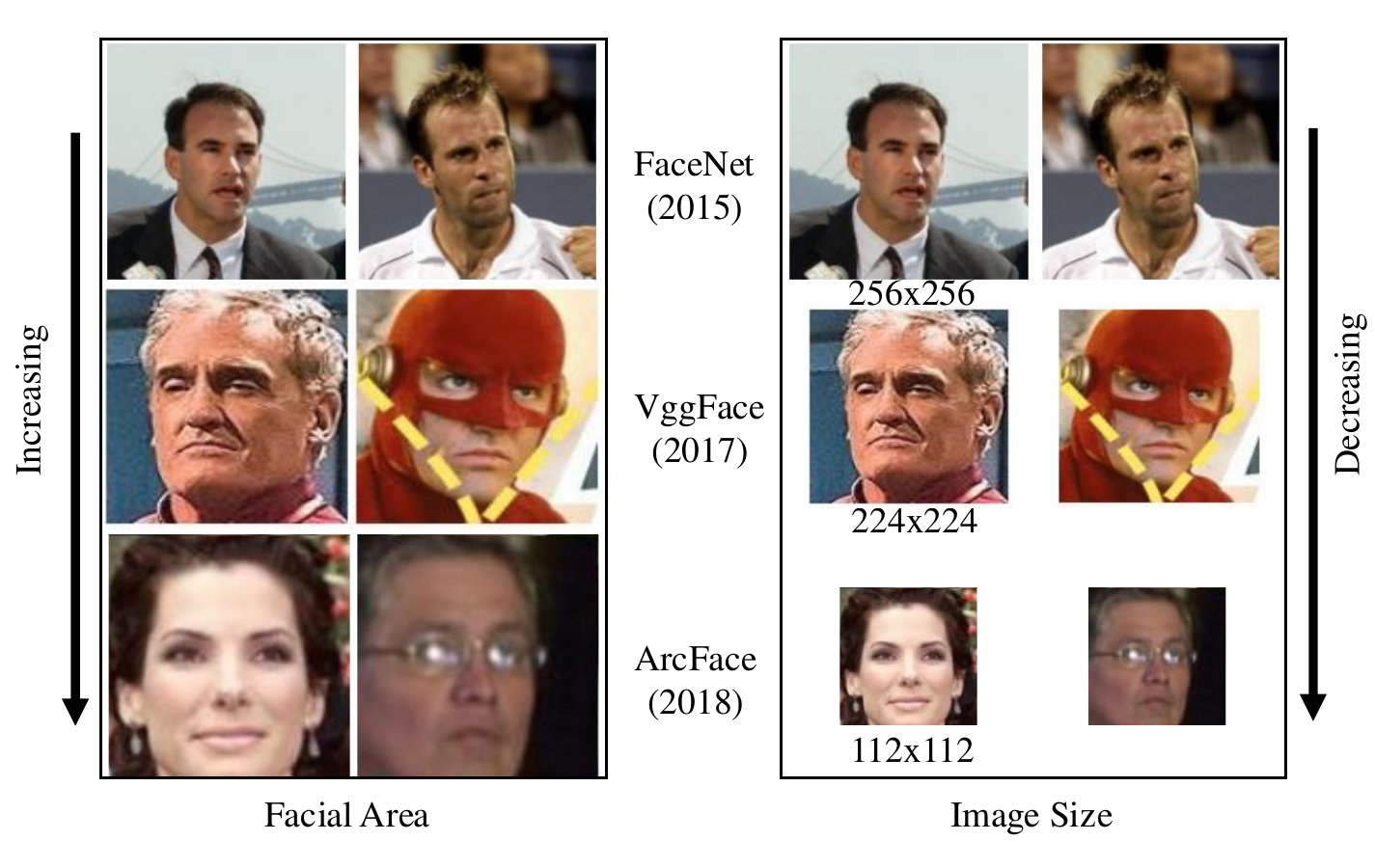}
    \vspace{-2mm}
    \caption{Comparison of FR in terms of facial contextual area and image resolution. Models have evolved to focus on more tightly cropped regions while reducing resized input image size ($256\times 256$ to $112 \times112$), enabling more efficient feature extraction.}
    \label{fig:facesize}
    \vspace{-4mm}
\end{figure}

\noindent\textbf{Efficiency, Adaptation, and Compact Embeddings: } Beyond achieving maximum accuracy, research has also focused on developing efficient architectures suitable for deployment on resource-constrained devices like mobile phones.  MobileFaceNet~\cite{chen2018mobilefacenets} employs depthwise separable convolutions to significantly reduce computational cost and model size while maintaining reasonable accuracy (IResNet100 at 99.83\% vs MobileFaceNet at 99.55\% in LFW verification accuracy while being 60$\times$ smaller). S-ViT\cite{kim2023s} applies sparse attention to reduce computational cost without sacrificing accuracy. The continuous evolution of neural network architectures, from deeper CNNs to attention-based Transformers and efficient mobile designs, has been a key driver alongside loss function innovations and larger datasets in pushing the performance boundaries of automated FR.

Fine-tuning is crucial for adapting FR models to new domains, especially under quality mismatches~\cite{liu2024farsight} (\textit{e.g.} low vs high quality images). Instead of full fine-tuning, which risks catastrophic forgetting, recent work like PETALface~\cite{narayan2025petalface} uses LoRA~\cite{hu2022lora}, a parameter-efficient finetuning method that adds low-rank adaptation modules. By weighting LoRA blocks based on image quality, PETALface adapts effectively to low-resolution faces while preserving high-resolution performance.

\begin{table*}[t]
\centering
\small
\setlength{\tabcolsep}{4pt}
\caption{Comparison of synthetic face training datasets for FR across five standard benchmarks. "Gap to Real" shows the average performance drop relative to the use of real CASIA-WebFace dataset alone for training. Brackets in the Generator Training Dataset column denote datasets used for pretraining, which may help models learn facial priors. A more fair comparison might involve equalizing the use of pretrained models. FR model used is IR50~\cite{qiu2021synface,bae2023digiface,kim2023dcface}.}

\resizebox{\linewidth}{!}{

\begin{tabular}{lccccccccccc}
\toprule
\textbf{Methods} & \textbf{Venue} & \textbf{Generator Train Dataset} & \textbf{\# images (\# IDs$\times$ imgs/ID)} & \textbf{LFW} & \textbf{CFP-FP} & \textbf{CPLFW} & \textbf{AgeDB} & \textbf{CALFW} & \textbf{Avg} & \textbf{Gap to Real} \\
\midrule\midrule
SynFace~\cite{qiu2021synface} & ICCV21 & FFHQ~\cite{karras2019style} & 0.5M (10K $\times$ 50) & 91.93 & 75.03 & 70.43 & 61.63 & 74.73 & 74.75 & 26.04 \\
DigiFace~\cite{bae2023digiface} & WACV23 & 511 3D Scans & 1M (10K $\times$ 100) & 95.40 & 87.40 & 78.87 & 76.97 & 78.62 & 83.45 & 11.34 \\
DCFace~\cite{kim2023dcface} & CVPR23 & CASIA-WebFace (FFHQ) & 0.5M (10K $\times$ 50) & 98.55 & 85.33 & 82.62 & 89.70 & 91.60 & 89.56 & 5.23 \\
IDnet~\cite{kolf2023identity} & CVPR23 & CASIA-WebFace~\cite{casia} & 0.5M (10K $\times$ 50) & 92.58 & 75.40 & 74.25 & 63.88 & 79.90 & 79.13 & 15.66 \\
ExFaceGAN~\cite{boutros2023exfacegan} & IJCB23 & CASIA-WebFace & 0.5M (10K $\times$ 50) & 93.50 & 73.84 & 71.60 & 78.92 & 82.98 & 80.17 & 14.62 \\
SFace2~\cite{boutros2024sface2} & TBIS24 & CASIA-WebFace & 0.6M (10K $\times$ 60) & 96.50 & 77.11 & 74.60 & 77.37 & 83.40 & 81.62 & 13.17 \\
Arc2Face~\cite{papantoniou2024arc2face} & ECCV24 &  WF42M~\cite{webface} (Stable Diffusion) & 0.5M (10K $\times$ 50) & 98.81 & \textbf{91.87} & 85.16 & 90.18 & 92.63 & 91.73 & 3.06 \\
Vec2Face~\cite{wu2024vec2face} & ICLR2025 & CASIA-WebFace (WebFace4M) & 0.5M (10K $\times$ 50) & \textbf{98.87} & 88.97 & \textbf{85.47} & \textbf{93.12} & \textbf{93.57} & \textbf{92.00} & \textbf{2.79} \\
\midrule
CASIA-WebFace (Real) & - & NA & 0.49M (Real) & 99.38 & 96.91 & 89.78 & 94.50 & 93.35 & 94.79 & 0.00 \\
\bottomrule
\end{tabular}
}
\label{tab:synthetic_comparison}
\vspace{-2mm}
\end{table*}

Also, FaceNet~\cite{schroff2015facenet} indicates that for a given training dataset, higher performance was achieved with d=128 compared to d=512. This finding implies that standard high-dimensional face embeddings contain significant redundancy, suggesting the potential to develop much more compact templates that could enable faster search and more efficient storage while retaining discriminative power.

\textcolor{cred}{The optimal embedding dimensionality varies with the scale and diversity of the training data. The embedding space must fit all subjects within a hypersphere while preserving sufficient inter-class distances. As the number of identities increases, a higher-dimensional feature space may be needed to maintain discriminability. Hence, smaller dimensions (e.g., 128) may suffice for moderate datasets, whereas larger datasets benefit from higher-dimensional representations to preserve subject separation. \reviewtag{\Rone}}

\subsection{Synthetic Datasets}

The growing demand for large-scale, diverse, and ethically sourced training datasets has driven increasing interest in the use of \textit{synthetic face data}. Collecting real-world facial datasets often introduces privacy and consent challenges, as well as issues of demographic imbalance and limited representation under challenging conditions ({\it e.g.}, extreme poses, rare ethnicities). Synthetic data offers a compelling alternative or data augmentation by enabling controlled, scalable, and bias-aware dataset creation~\cite{qiu2021synface,kim2023dcface,bae2023digiface}.

The generation of photorealistic synthetic faces has been significantly advanced by deep generative models, particularly Generative Adversarial Networks (GANs)\cite{goodfellow2014generative}. Variants such as StyleGAN\cite{karras2019style,Karras2019stylegan2} are especially effective at producing high-resolution facial imagery, capable of modeling complex visual distributions and allowing fine-grained control over attributes like pose, expression, and illumination via manipulations in the latent space. The integration of 3D Morphable Models (3DMM) into GANs has further enhanced the controllability of facial attributes during generation~\cite{deng2018uv,gecer2018semi,geng20193d,kim2018deep}. For instance, CFSM~\cite{liu2022controllable} leverages GANs to synthesize faces with diverse styles, aiding in the generation of richly varied datasets.

\begin{figure}[t!]
    \centering
    \includegraphics[width=0.8\linewidth]{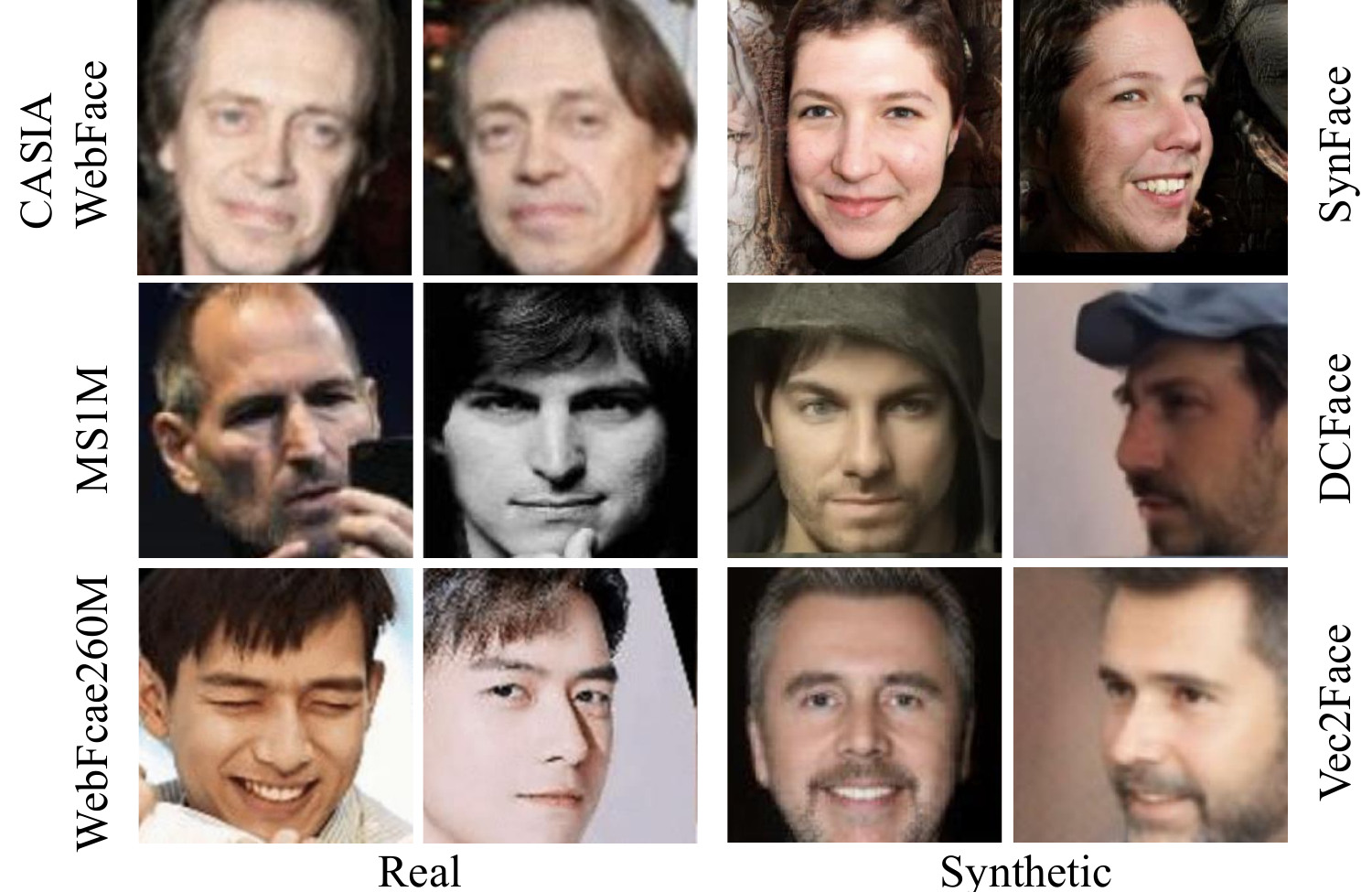}
    \caption{Examples of real and synthetic datasets. Real datasets evolve from the primarily frontal CASIA-WebFace to the large-scale, diverse WebFace260M,  including wide variations in pose and expression. Synthetic datasets also advance, focusing on improving diversity and maintaining identity consistency.}
    \label{fig:example_syn}
    \vspace{-4mm}
\end{figure}

Recently, Diffusion Models~\cite{ho2020denoising} emerge as a powerful generative paradigm, achieving impressive image quality and diversity. 
They generate images by gradually denoising a sample from pure noise, learning to reverse the diffusion process. 
Text-conditioned diffusion models are especially effective for controlled synthesis, enabling detailed and semantically guided generation~\cite{dhariwal2021diffusion,rombach2022high}.  
ControlNet~\cite{controlnet} and IP-Adapter~\cite{ipadapter} make the model adhere to input conditions such as facial landmarks, masks or other clues. 

Leveraging these generative capabilities, researchers have explored creating entire synthetic datasets specifically for training FR models. The goal of face dataset generation is to create multiple images of the same subject at a large scale. The ID consistency is at an interplay with the diversity of generated images.  SynFace~\cite{qiu2021synface} first applies GAN and latent interpolation method to generate face datasets, resulting in average face verification rate of $74.75\%$, marking a significant drop compared to real CASIA-Webface dataset of $94.79\%$ as in Tab.~\ref{tab:synthetic_comparison}. 
Since then, mulitple works have attempted to reduce the gap. 

Following SynFace, rapid advances have sought to bridge the gap between synthetic and real face datasets. Diffusion-based models such as DCFace~\cite{kim2023dcface} separate identity and style conditions to produce identity-consistent and diverse subjects, while Arc2Face~\cite{papantoniou2024arc2face} builds on pretrained Stable Diffusion to exploit the generalization power of foundation models. Vec2Face~\cite{wu2024vec2face} shows that GAN-based synthesis remains competitive when guided by a FR feature space, underscoring the importance of identity disentanglement. These methods exemplify the trend toward improving realism, diversity, and identity consistency to narrow the gap with real data. Tab.~\ref{tab:synthetic_comparison} presents the performance of recent synthetic face datasets on FR.

There is growing interest in synthetic FR challenges. FRCSyn series~\cite{melzi2024frcsyn,deandres2024frcsyn}, show that while synthetic-only training trails real data slightly, it reduces demographic bias and improves robustness to pose, age, and occlusion. Figure~\ref{fig:example_syn} compares real and synthetic datasets. Adding synthetic data, such as Vec2Face~\cite{wu2024vec2face} to CASIA-WebFace, can raise average verification accuracy by about 1.00\%.

Despite its promise, using purely synthetic data to train SoTA FR models faces key challenges, primarily the \textit{domain gap} between synthetic and real images. Models trained only on synthetic data often struggle to generalize to real-world due to subtle differences in texture, lighting, or artifacts from the generation process. Achieving sufficient diversity and realism, especially in capturing identity nuances under varying conditions, remains an active research focus. 
\textcolor{cred}{Generative models can mitigate privacy and consent concerns, yet their training on real web-sourced images raises doubts about whether generated content removes or only obscures data ownership and consent issues~\cite{Silberling2025}. To mitigate these risks, growing attention is given to watermarking and related methods~\cite{asnani2022proactive} that clearly mark images as synthetic. \reviewtag{\Rone}
}

\subsection{Feature Fusion in Face Recognition}

In template-based FR, multiple face images of the same identity—often captured under varying conditions of pose, illumination, resolution, and occlusion—must be fused into a single, compact representation to enable efficient and accurate comparison. This fusion scenario commonly arises in gallery settings, where multiple still images (or media) per subject must be aggregated into a unified template.

A second, and increasingly popular, scenario involves video-based FR, where frames extracted from probe video sequences are fused into a single representation. This use case poses unique challenges, as it often requires online (on-the-fly) feature fusion to support real-time applications such as surveillance or mobile authentication. Despite differing temporal constraints, both still-image and video-based fusion share the core objective: to generate robust and compact representations that preserve discriminative identity cues.

Feature fusion is a critical step in this process, as it determines how the information from diverse images of the same person is aggregated into a unified descriptor. Naive methods like average or max pooling treat all feature embeddings equally, which can dilute discriminative cues by giving equal importance to low-quality or redundant images. Effective feature fusion must not only compress but also intelligently filter, weight, and adapt to the content of the input set. The ability to generate order-invariant, and compact template representations directly impacts FR performance, especially in unconstrained or real-time scenarios.

Early video-based FR approaches employed adaptive hidden Markov models to capture temporal dynamics and recognize entire video sequences~\cite{tu2009optimizing}. Over time, feature fusion in FR advanced from simple averaging to adaptive and context-aware neural aggregation. The Neural Aggregation Network (NAN)~\cite{yang2017neural} demonstrated the effectiveness of learning quality-aware attention weights for robust, order invariant face templates. Building on this, Multicolumn Networks~\cite{xie2018multicolumn} and C-FAN~\cite{gong2019video} introduced fine-grained quality analysis by modeling visual and contextual importance or by weighting individual feature channels. These developments significantly improved FR performance on challenging template-based benchmarks such as IJB-C~\cite{ijbc}.

Recent work emphasizes scalable and generalized feature fusion across diverse conditions. Methods like CAFace~\cite{kim2022cluster}, CoNAN~\cite{jawade2023conan}, and ProxyFusion~\cite{jawade2024proxyfusion} sustain performance even with templates containing many varied images. Practical approaches such as Norm Pooling~\cite{nanduri2024template} show that simple heuristics can be effective in multi domain settings, especially with limited training data. Overall, these innovations reflect a trend toward scalable, efficient strategies that handle long videos while maintaining robustness.

\section{State of the art in Face Recognition}

\textcolor{cred}{FR evaluation progressed from lab specific testing to standardized protocols~\cite{phillips1998feret,phillips2000feret,phillips2005overview}. Before deep learning, small proprietary datasets with fewer than 100 subjects and over 95\% reported accuracy lacked generalizability~\cite{turk1991face,belhumeur1997eigenfaces,martinez2001pca}. The 1996 FERET program established large scale standardized evaluation, enabling systematic benchmarking and transparent comparison~\cite{phillips1998feret,phillips2000feret}. This shift to structured evaluation allowed consistent progress tracking and set stage for deep learning’s transformative impact on FR~\cite{krizhevsky2012imagenet,taigman2014deepface,schroff2015facenet,guo2019survey,wang2021deep,wang2022survey}. 
}

\textcolor{cred}{The FERET program~\cite{phillips1998feret} in the 1990s standardized FR benchmarks with 14{,}126 images of 1{,}199 individuals. A decade later, the FRGC~\cite{phillips2005overview} expanded this with 50{,}000 high resolution images and 3D scans, defining protocols that guided later NIST FRVT~\cite{grother2019face} series. Together, they formed the basis of modern FR benchmarks. \reviewtag{\Rthree, \Rfour}
}






\begin{table*}[htbp]
\centering
\begin{minipage}[t]{0.48\textwidth}
\centering
\tiny
\caption{Performance on CFP-FP~\cite{cfpfp} Dataset}
\begin{tabular}{|l|l|l|l|c|}
\hline
\textbf{Method Name} & \textbf{Backbone} & \textbf{Loss Function} & \textbf{Training Data} & \textbf{Verification (\%)} \\
\hline
GFace~\cite{zhao2025global} & IResNet-50 & GCE (LO) & Casia-WebFace & 97.44 \\
CosFace~\cite{wang2018cosface} & ResNet100 & CosFace~\cite{wang2018cosface} & MS1MV2 & 98.13 \\
ArcFace~\cite{deng2019arcface} & ResNet101 & ArcFace & MS1MV2 & 98.27 \\
MV-Softmax~\cite{wang2020mis} & ResNet100 & MV-Softmax & MS1MV2 & 98.28 \\
CurricularFace~\cite{huang2020curricularface} & ResNet101 & CurricularFace & MS1MV2 & 98.37 \\
TransFace-B~\cite{dan2023transface} & ResNet100 & ArcFace & MS1MV2 & 98.39 \\
MagFace~\cite{meng2021magface} & ResNet100 & MagFace & MS1MV2 & 98.46 \\
AdaFace~\cite{kim2022adaface} & ResNet101 & AdaFace & MS1MV2 & 98.49 \\
CQA-Face~\cite{wang2022cqa} & ResNet100 & CQA-Face & MS1MV2 & 98.49 \\
UniFace~\cite{zhou2023uniface} & ResNet100 & UniFace~\cite{zhou2023uniface}  & MS1MV2 & 98.63 \\
URL~\cite{shi2020towards} & ResNet101 & URL & MS1MV2 & 98.64 \\
LGAF~\cite{wang2025local} & ResNet100 & ArcFace & MS1MV2 & 98.77 \\
ArcFace~\cite{deng2019arcface} & ResNet50 & ArcFace & Glint360K & 98.77 \\
ViT-S~\cite{dosovitskiy2020image} & ViT-S & ArcFace & Glint360K & 98.85 \\
CosFace + KP-RPE & ViT & CosFace & WebFace4M & 98.91 \\
TransFace-S~\cite{dan2023transface} & ViT-S & ArcFace & Glint360K & 98.91 \\
AdaFace~\cite{kim2022adaface} & ViT & AdaFace & WebFace4M & 98.94 \\
KP-RPE~\cite{kim2024keypoint} & ViT & AdaFace & WebFace4M & 99.01 \\
ViT-B~\cite{dosovitskiy2020image} & ViT-B & ArcFace & Glint360K & 99.02 \\
AdaFace~\cite{kim2022adaface} & ResNet101 & AdaFace & MS1MV3 & 99.03 \\
R100 & ResNet100 & ArcFace & Glint360K & 99.04 \\
AdaFace~\cite{kim2022adaface} & ViT & AdaFace & MS1MV3 & 99.06 \\
ArcFace~\cite{deng2019arcface} & ResNet101 & ArcFace & WebFace4M & 99.06 \\
KP-RPE~\cite{kim2024keypoint} & ViT & ArcFace & WebFace4M & 99.09 \\
ViT-L & ViT-L & ArcFace & Glint360K & 99.10 \\
KP-RP~\cite{kim2024keypoint}E & ViT & AdaFace & MS1MV3 & 99.11 \\
GFace~\cite{zhao2025global} & IResNet-100 & GCE (LO) & MS1MV3 & 99.12 \\
R200 & ResNet200 & ArcFace & Glint360K & 99.14 \\
AdaFace~\cite{kim2022adaface} & ResNet101 & AdaFace & WebFace4M & 99.17 \\
TransFace-B~\cite{dan2023transface} & ViT-B & ArcFace & Glint360K & 99.17 \\
AdaFace~\cite{kim2022adaface} & ResNet101 & AdaFace & WebFace12M & 99.24 \\
KP-RPE~\cite{kim2024keypoint} & ViT & AdaFace & WebFace12M & \textbf{99.30} \\
TransFace-L~\cite{dan2023transface} & ViT-L & ArcFace & Glint360K & \textbf{99.32} \\
\hline
\end{tabular}
\label{tab:sota1}
\end{minipage}
\hfill
\begin{minipage}[t]{0.48\textwidth}
\centering
\tiny
\caption{Performance on IJB-C~\cite{ijbc} Dataset}
\begin{tabular}{|l|l|l|l|c|}
\hline
\textbf{Method Name} & \textbf{Backbone} & \textbf{Loss Function} & \textbf{Training Data} & \textbf{TAR@FAR=1e-4} \\
\hline
ArcFace~\cite{deng2019arcface} & ResNet101 & ArcFace~\cite{deng2019arcface} & MS1MV2 & 96.03 \\
MagFace~\cite{meng2021magface} & ResNet101 & MagFace~\cite{meng2021magface} & MS1MV2 & 95.81 \\
MagFace+IIC & ResNet101 & MagFace & MS1MV2 & 95.89 \\
ViT-S & ViT-S & ArcFace & MS1MV2 & 95.89 \\
CurricularFace~\cite{huang2020curricularface} & ResNet101 & CurricularFace & MS1MV2 & 96.10 \\
ViT-B & ViT-B & ArcFace & MS1MV2 & 96.15 \\
ViT-L & ViT-L & ArcFace & MS1MV2 & 96.24 \\
TransFace-S~\cite{dan2023transface} & ViT-S & ArcFace & MS1MV2 & 96.45 \\
TransFace-B~\cite{dan2023transface} & ViT-B & ArcFace & MS1MV2 & 96.55 \\
TransFace-L~\cite{dan2023transface} & ViT-L & ArcFace & MS1MV2 & 96.59 \\
ArcFace+CFSM & ResNet101 & ArcFace & MS1MV2 & 96.60 \\
ARoFace~\cite{saadabadi2024aroface} & ResNet101 & ArcFace & MS1MV2 & 96.66 \\
ElasticFace~\cite{boutros2022elasticface} & ResNet101 & ElasticFace & MS1MV2 & 96.65 \\
TopoFR~\cite{dan2024topofr} & ResNet101 & TopoFR & MS1MV2 & 96.95 \\
GFace~\cite{zhao2025global} & ResNet101 & TopoFR & MS1MV2 & 96.96 \\
AdaFace~\cite{kim2022adaface} & ResNet101 & AdaFace & MS1MV2 & 97.09 \\
KP-RPE~\cite{kim2024keypoint} & ViT-B & CosFace & WebFace4M & 96.98 \\
TopoFR~\cite{dan2024topofr} & ResNet200 & TopoFR & MS1MV2 & 97.08 \\
AdaFace~\cite{kim2022adaface} & ViT-B & AdaFace & MS1MV3 & 97.10 \\
KP-RPE~\cite{kim2024keypoint} & ViT-B & AdaFace & WebFace4M & 97.13 \\
AdaFace~\cite{kim2022adaface} & ViT-B & AdaFace & WebFace4M & 97.14 \\
KP-RPE~\cite{kim2024keypoint} & ViT-B & AdaFace & MS1MV3 & 97.16 \\
KP-RPE~\cite{kim2024keypoint} & ViT-B & ArcFace & WebFace4M & 97.21 \\
PartialFC~\cite{an2021partial} & ResNet101 & ArcFace & WebFace4M & 97.22 \\
CatFace~\cite{talemi2024catface} & ResNet101 & CatFace & MS1MV2 & 97.43 \\
AdaFace~\cite{kim2022adaface} & ResNet101 & AdaFace & WebFace4M & 97.39 \\
ARoFace~\cite{saadabadi2024aroface} & ResNet101 & AdaFace & WebFace4M & 97.51 \\
AdaFace~\cite{kim2022adaface} & ResNet101 & AdaFace & WebFace12M & 97.66 \\
PartialFC~\cite{an2021partial} & ResNet101 & ArcFace & WebFace12M & 97.58 \\
ARoFace~\cite{saadabadi2024aroface} & ResNet101 & AdaFace & WebFace12M & 97.60 \\
TopoFR~\cite{dan2024topofr} & ResNet101 & TopoFR & Glint360K & 97.60 \\
KP-RPE~\cite{kim2024keypoint} & ViT-B & AdaFace & WebFace12M & 97.82 \\
PartialFC~\cite{an2021partial} & ResNet101 & ArcFace & WebFace42M & 97.82 \\
TopoFR~\cite{dan2024topofr} & ResNet200 & TopoFR & Glint360K & 97.84 \\
PartialFC~\cite{an2021partial} & ViT-B & ArcFace & WebFace42M & 97.90 \\
PartialFC~\cite{an2021partial} & ResNet200 & ArcFace & WebFace42M &\textbf{ 97.97} \\
UniTSFace~\cite{jia2023unitsface} & ViT-L & UniTSFace & WebFace42M & \textbf{97.99} \\
\hline
\end{tabular}
\label{tab:sota2}
\end{minipage}
\\
\begin{minipage}[t]{0.48\textwidth}
\centering
\tiny
\caption{Performance on IJB-S~\cite{ijbs} Dataset}
\begin{tabular}{|l|l|l|l|c|c|}
\hline
\textbf{Method Name} & \textbf{Backbone} & \textbf{Loss Function} & \textbf{Training Data} & \textbf{Rank-1} & \textbf{Rank-5} \\
\hline
PFE~\cite{shi2019probabilistic} & ResNet101 & PFE & MS1MV2 & 50.16 & 58.33 \\
URL~\cite{shi2020towards} & ResNet101 & URL & MS1MV2 & 59.79 & 65.78 \\
CurricularFace~\cite{huang2020curricularface} & ResNet101 & CurricularFace & MS1MV2 & 62.43 & 68.68 \\
AdaFace~\cite{kim2022adaface} & ResNet101 & AdaFace & MS1MV2 & 65.26 & 70.53 \\
AdaFace~\cite{kim2022adaface} & ViT & AdaFace & MS1MV3 & 65.95 & 71.64 \\
AdaFace~\cite{kim2022adaface} & ResNet101 & AdaFace & MS1MV3 & 67.12 & 72.67 \\
 KP-RPE~\cite{kim2024keypoint} & ViT & AdaFace & MS1MV3 & 67.62 & 73.25 \\
ArcFace~\cite{deng2019arcface} & ResNet101 & ArcFace & WebFace4M & 69.26 & 74.31 \\
AdaFace~\cite{kim2022adaface} & ResNet101 & AdaFace & WebFace4M & 70.42 & 75.29 \\
ARoFace~\cite{saadabadi2024aroface} & ResNet101 & ArcFace & WebFace4M & 70.96 & 75.54 \\
AdaFace~\cite{kim2022adaface} & ResNet101 & AdaFace & WebFace12M & 71.35 & 76.24 \\
AdaFace~\cite{kim2022adaface} & ViT & AdaFace & WebFace4M & 71.90 & 77.09 \\
 KP-RPE~\cite{kim2024keypoint} & ViT & CosFace & WebFace4M & 72.22 & 77.67 \\
ARoFace~\cite{saadabadi2024aroface} & ResNet101 & AdaFace & WebFace12M & 72.28 & 77.93 \\
 KP-RPE~\cite{kim2024keypoint} & ViT & AdaFace & WebFace4M & \textbf{72.78} & \textbf{78.20 }\\
KP-RPE~\cite{kim2024keypoint} & ViT & ArcFace & WebFace4M & \textbf{73.04 }& \textbf{78.62} \\
\hline
\end{tabular}
\label{tab:sota4}
\end{minipage}
\hfill
\begin{minipage}[t]{0.48\textwidth}
\vspace{0.43cm}
\centering
\tiny
\caption{Performance on TinyFace~\cite{cheng2018low} Dataset}
\begin{tabular}{|l|l|l|l|c|c|}
\hline
\textbf{Method Name} & \textbf{Backbone} & \textbf{Loss Function} & \textbf{Training Data} & \textbf{Rank1} & \textbf{Rank5} \\
\hline
ArcFace+CFSM & ResNet101 & ArcFace & MS1MV2 & 64.69 & 68.80 \\
TransFace-L~\cite{dan2023transface} & ViT-S & ArcFace & MS1MV2 & 67.52 & 71.00 \\
ARoFace~\cite{saadabadi2024aroface} & ResNet101 & ArcFace & MS1MV3 & 67.54 & 71.05 \\
LGAF~\cite{wang2025local} & ResNet101 & ArcFace & MS1MV2 & 68.35 & 71.59 \\
ArcFace~\cite{deng2019arcface} & ResNet101 & ArcFace & WebFace4M & 71.11 & 74.38 \\
AdaFace~\cite{kim2022adaface} & ResNet101 & AdaFace & WebFace4M & 72.02 & 74.52 \\
AdaFace~\cite{kim2022adaface} & ResNet101 & AdaFace & WebFace12M & 72.29 & 74.97 \\
REE~\cite{chai2023recognizability} & ResNet-50 & ArcFace & Native VLR & 73.06 & 77.22 \\
ARoFace~\cite{saadabadi2024aroface} & ResNet101 & ArcFace & WebFace4M & 73.80 & 76.53 \\
ARoFace~\cite{saadabadi2024aroface} & ResNet101 & AdaFace & WebFace4M & 73.98 & 76.47 \\
ARoFace~\cite{saadabadi2024aroface} & ResNet101 & AdaFace & WebFace12M & 74.00 & 76.87 \\
 KP-RPE~\cite{kim2024keypoint} & ViT-B & CosFace & WebFace4M & 75.48 & 78.30 \\
 KP-RPE~\cite{kim2024keypoint} & ViT-B & ArcFace & WebFace4M & \textbf{75.62} & \textbf{78.57} \\
 KP-RPE~\cite{kim2024keypoint} & ViT-B & AdaFace & WebFace4M & \textbf{75.80} & \textbf{78.49} \\
\hline
\end{tabular}
\label{tab:sota3}
\end{minipage}

\end{table*}

\subsection{Benchmark Evaluations}

Robust evaluation of modern FR systems necessitates standardized benchmark datasets that reflect various real-world challenges. Prominent evaluation datasets extensively cited in recent literature include Labeled Faces in the Wild (LFW)~\cite{lfw}, CFP-FP~\cite{cfpfp}, CP-LFW~\cite{cplfw}, AgeDB~\cite{agedb}, YouTube Faces (YTF)~\cite{wolf2011face}, TinyFace~\cite{cheng2018low}, and several iterations of the IARPA Janus Benchmark (IJB) series such as IJB-B~\cite{ijbb}, IJB-C~\cite{ijbc}, and IJB-S~\cite{ijbs}. Each dataset addresses specific challenges inherent in FR scenarios. Below some datasets are described in detail.

\noindent\textbf{Labeled Faces in the Wild (LFW~\cite{lfw})} consists of over 13,000 facial images collected from the web, annotated with identity labels. The dataset includes multiple images for approximately 1,680 individuals of high-quality images. Main usage of this dataset is for the verification task. 

\noindent\textbf{YouTube Faces (YTF~\cite{wolf2011face})} specifically targets video-based unconstrained FR. The dataset comprises clips varying from 48 to 6,070 frames, with an average length of 181 frames and contains an average of 2 videos per subject, making it useful for assessing algorithms designed to handle real-world variability in videos.

\noindent\textbf{CFP-FP~\cite{cfpfp}} evaluates the capability of algorithms to match frontal face images with their corresponding profile ones. 
It is particularly challenging due to large variations in facial orientation. 
The dataset is widely used to benchmark algorithms designed for pose-invariant face verification.

\noindent\textbf{TinyFace~\cite{cheng2018low}} is explicitly designed for low-resolution FR research at scale. It includes 169,403 naturally low-resolution images (average size 20x16 pixels) depicting 5,139 identities. Images in TinyFace are cropped from crowded scenes and span a diverse range of lighting, occlusion, backgrounds.

\noindent\textbf{IARPA Janus Benchmark-C (IJB-C)} expands upon earlier series IJB-B~\cite{ijbb}, containing imagery and videos for 3,531 subjects, including 1,661 newly added identities. It comprises approximately 138,000 images and 11,000 videos. IJB-C serves as a challenging dataset for template-based recognition tasks. It contains significant variations in pose, illumination, and image quality~\cite{ijbc}. Performance is often reported as TAR@FAR=threshold where the FAR threshold is selected based on the target operating point.

\begin{figure*}[t!]
    \centering
    \includegraphics[width=\linewidth]{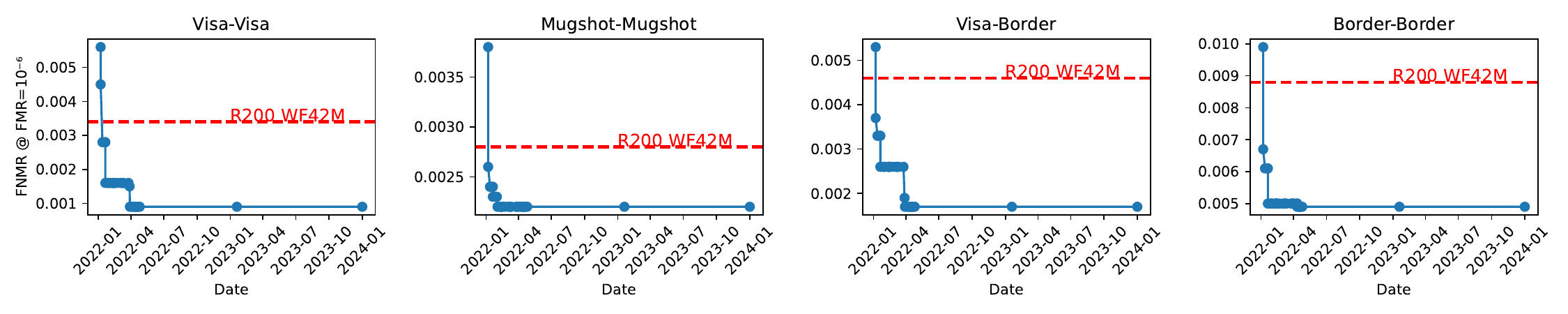}
    \vspace{-6mm}
\caption{SoTA performance in NIST FRVT 1:1 verification since January 2022. Plots show the cumulative minimum False Non-Match Rate (FNMR) achieved by any submitted algorithm up to the corresponding date for the Visa, Mugshot, Visa Border, and Border datasets (the plot titles indicate gallery - probe in order). Performance shows a low False Match Rate ({\it e.g.}, FMR=$10^{-6}$). The dashed red line indicates the performance level achieved by the WebFace42M entry (R200~\cite{webface} WF42M) for comparison.}
    \label{fig:frvt}
    \vspace{-4mm}
\end{figure*}

\noindent\textbf{IARPA Janus Surveillance Video Benchmark (IJB-S)} targets surveillance-specific scenarios, featuring images and videos of 202 subjects collected at a Department of Defense facility. The galleries are comprised of high-quality upper torso images and the probes are videos captured by surveillance camera of varied altitude and range. It is suited for evaluating surveillance-oriented FR approaches~\cite{ijbs}.

Collectively, these datasets represent comprehensive benchmarks that drive progress in addressing the nuanced challenges of modern FR technologies.

\subsubsection{State-of-the-Art Performance}
FR performance shows progress across different benchmarks, underscoring the effectiveness of deep learning architectures and sophisticated loss functions. In Tab.~\ref{tab:sota1}$\sim$\ref{tab:sota4} we compile the up-to-date FR performance under each evaluation datasets using relevant metrics.

In 1:1 verification, Verification Accuracy is the proportion of correct matches and nonmatches. On difficult datasets, performance is given by the True Accept Rate (TAR) at a fixed False Accept Rate (FAR), usually FAR=0.01\%, representing $1 - \text{FNMR}$ at a fixed FMR. For 1:N identification, Rank-$k$ accuracy shows how often the correct identity is within the top $k$ results (Rank-1 is strictest). In open set identification, the True Positive Identification Rate (TPIR) at a given False Positive Identification Rate (FPIR), such as FPIR=0.01\%, measures correct identifications while limiting false matches of unknowns.

\noindent\textbf{CFP-FP~\cite{cfpfp}:}
 Current methods achieve extremely high verification accuracy, often exceeding 99\%. Top performance is typically seen with models utilizing ViT~\cite{dosovitskiy2020image} backbones ({\it e.g.}, ViT-L, ViT-B, ViT-S variants) or deeper ResNet~\cite{he2016deep} architectures ({\it e.g.}, ResNet-101, ResNet-200). Effective loss functions like AdaFace~\cite{kim2022adaface} and ArcFace~\cite{deng2019arcface} are prevalent among the leading methods. Furthermore, training on very large datasets like Glint360K~\cite{an2021partial}, WebFace~\cite{webface} is crucial for reaching the highest scores, with methods like TransFace~\cite{dan2023transface} reporting accuracies above 99.3\%.

\noindent\textbf{IJB-C~\cite{ijbc}:} IJB-C dataset presents a more challenging scenario involving template-based matching (comparing sets of images/video frames). Performance is often measured by the TAR@FAR=0.01\%. SoTA methods, such as PFC~\cite{an2021partial} (utilizing ViT-L or ResNet200), KP-RPE~\cite{kim2024keypoint} (with ViT-B), AdaFace~\cite{kim2022adaface}, and TopoFR~\cite{dan2024topofr}, achieve TAR values around 98\% at 0.01\%. Again, larger backbones (ViT-L~\cite{dosovitskiy2020image}, ResNet200~\cite{he2016deep}) and extensive training data (WebFace42M~\cite{webface}, Glint360K~\cite{an2021partial}) are characteristic of the top-performing approaches.

\noindent\textbf{TinyFace~\cite{cheng2018low}} (Low-Resolution Recognition): TinyFace specifically addresses the difficulty of recognizing faces from very low-resolution images. As expected, performance metrics like Rank-1 identification accuracy are considerably lower than on high-resolution datasets. Leading methods, predominantly using ViT-B backbones combined with techniques like KP-RPE that make the model robust to misalignments and loss functions such as AdaFace that allow quality adaptive training achieve Rank-1 accuracies around 75-76\%. Training on large datasets like WebFace12M is also helpful for performance. Methods like ARoFace~\cite{saadabadi2024aroface} also show competitive results, highlighting the ongoing efforts to improve recognition under significant resolution constraints.

\noindent\textbf{IJB-S~\cite{ijbs}}: Similar to TinyFace, IJB-S contains low-quality imageries and presents faces extracted from surveillance footage. We report Surveillance-to-Still (S2S) protocol. Top performance, measured by S2S Rank-1 accuracy, reaches approximately 73\%.  Another characteristic of this dataset is that the template size is large, making it a suitable choice to evaluate the methods for template feature fusion methods~\cite{kim2022cluster,jawade2024proxyfusion,jawade2023conan}.

\subsection{Technology Evaluations by NIST}

NIST has been conducting independent evaluations of FR technologies since 1999. Initially under the Face Recognition Vendor Test (FRVT), these activities have expanded to include the Face Recognition Technology Evaluation (FRTE) and Face Analysis Technology Evaluation (FATE)~\cite{nist_frte_fate}. NIST evaluations are critical for assessing the readiness of algorithms for real-world deployment.

Unlike benchmark evaluations conducted on public datasets, NIST uses sensitive operational datasets not available in the public domain, such as mugshots, visa application photos, and imagery from border kiosks. Developers submit their algorithms for third-party testing, ensuring unbiased, standardized evaluation. NIST assesses both 1:1 verification and 1:N identification scenarios, measuring metrics such as False Non-Match Rate (FNMR) at a fixed False Match Rate (FMR), and identification rates at various thresholds. NIST’s reporting of 1-FNMR at a fixed FMR is equivalent to the more recently used terminology TAR@FAR metric. The number of distinct algorithms submitted to FRVT has grown over time, reflecting its increasing relevance and accessibility. Since evaluation is free and ongoing, participants can submit algorithms at any time for both 1:1 verification and 1:N identification tasks, with N now including up to 12 million enrolled identities.

Between 2014 and 2018, NIST reported that FR software improved by a factor of 20 in search accuracy~\cite{nist_2018_progress}, highlighting the rapid pace of advancement in the field. To date, NIST has evaluated over 400 algorithms~\cite{grother2019face}. While academic benchmarks are crucial for driving research, the NIST FRVT provides an ongoing, operational evaluation of both academic and commercial algorithms under various scenarios, serving as a key indicator of the absolute SoTA deployed in real-world systems. Fig.~\ref{fig:frvt} plots recent FRVT 1:1 verification performance results, including the performance trajectory of a strong academic baseline (ResNet-200 trained on the large-scale WebFace42M dataset, highlighted in red) for comparison against numerous vendor submissions. The performance results from the FRVT evaluations are publicly available on the organizer’s website~\cite{grother2019face}.

The results consistently show that top-performing algorithms, often developed by industry players, achieve excellent accuracy. However, many leading academic models, especially those trained on large public datasets like WebFace42M, perform competitively, demonstrating the strong impact of academic research on real-world applications. Nonetheless, the very best performing systems typically originate from industry, a difference that may stem from access to larger proprietary datasets, specialized hardware optimizations, extensive system-level engineering, or specific algorithmic refinements not yet published in academic literature. Still, the close proximity of top academic results to industrial leaders underscores the significant contribution of academic research to advancing practical FR capabilities.


\section{Current Challenges in Face Recognition}
\label{sec:challenges}

\begin{figure}[t]
    \centering
    \vspace{-3mm}
    \includegraphics[width=0.8\linewidth]{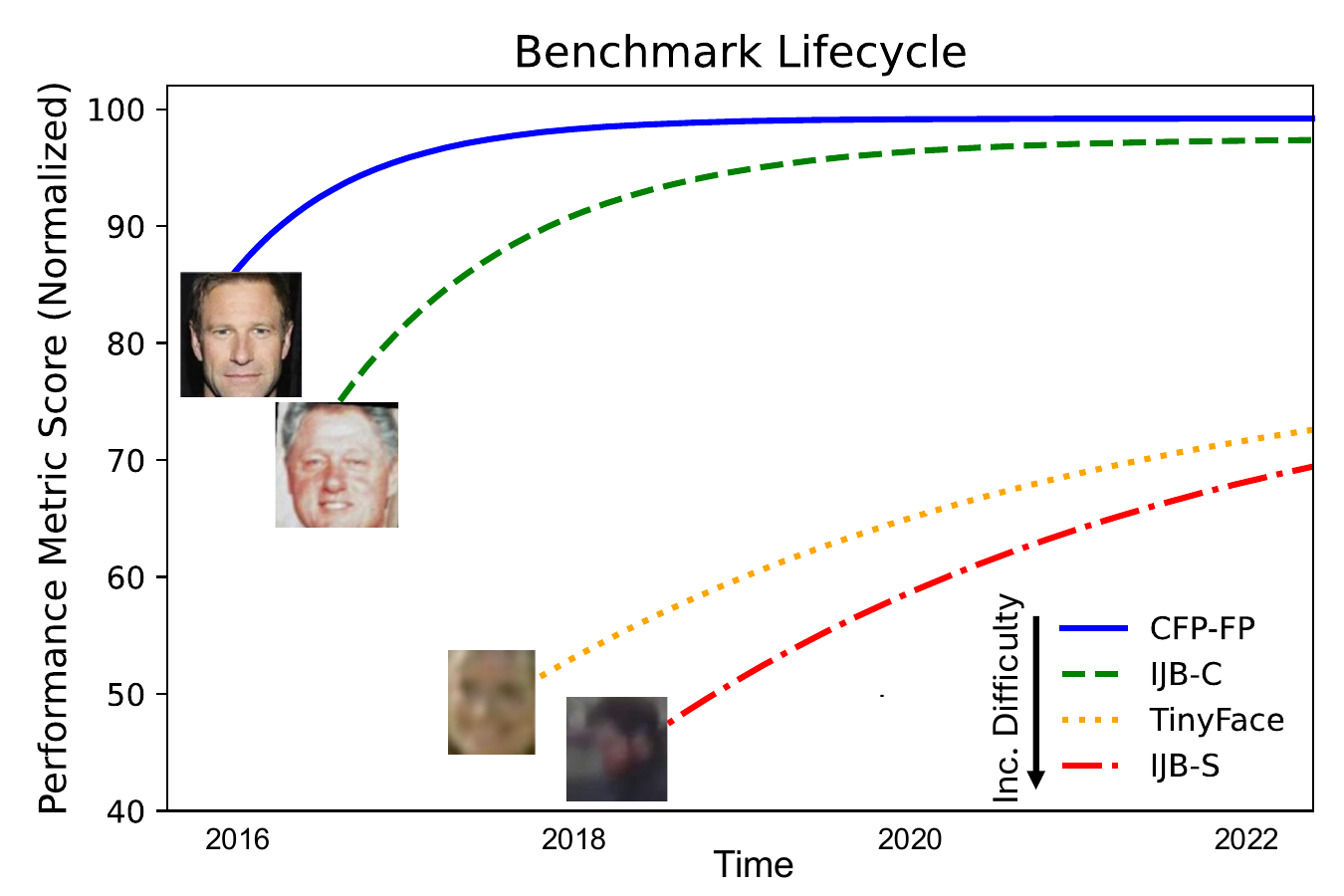}
\caption{Illustration of performance trends of various FR benchmarks over time, illustrating eventual performance saturation as datasets become extensively explored. This saturation is an indication of the progress in the field. }
    \label{fig:dataset_difficulty}
    \vspace{-4mm}
\end{figure}

Although FR performance has advanced greatly, challenges persist in real-world use. As Fig.~\ref{fig:dataset_difficulty} shows, benchmark saturation reflects limits in current evaluations rather than problem resolution. Fig.~\ref{fig:face_sim} illustrates that for low-quality images (\textit{e.g.}, IJB-S), missing facial details force models to depend on soft biometric cues like beards or hair color.

\subsection{Recognition at Scale: Large Galleries}
A major limitation of current academic benchmarks is their failure to match the scale and complexity of real biometric systems. Practical FR deployments often manage galleries with millions to billions of identities, far beyond the thousands used in research. For example, India's Aadhaar system stores biometric data, including facial images, for over 1.4 billion people. At such scales, even small drops in recognition accuracy lead to large numbers of false matches or misses, affecting millions of users.

An experiment on the IJB S dataset was conducted to illustrate the gap between benchmark evaluations and real world scenarios. The baseline IJB S gallery (202 identities) was expanded with external imposters ranging from 1,000 to 10,000 identities. Tab.~\ref{tab:ijbs_imposter_gallery} shows a performance drop as gallery size increases, revealing reduced recognition accuracy with larger galleries. This demonstrates that benchmarks can fail to represent real world conditions involving extensive galleries with diverse, unstructured, and noisy identities.

\begingroup
\color{cred}

\subsection{Practical Applications}

While benchmark performance provides a useful measure of algorithmic progress, the ultimate test of FR lies in its deployment across real-world applications. Each application domain introduces unique constraints, ethical considerations and  requirements that continue to shape FR research.

\noindent \textbf{Surveillance and Security.} 
Surveillance applications revealed limits of low-quality, cross-resolution FR, prompting benchmarks like IJB-S~\cite{ijbs} and TinyFace~\cite{cheng2018low} and quality-aware models~\cite{meng2021magface,kim2022adaface}. More recent BRIAR project~\cite{BRIAR} extends evaluation up to 1,000\,m standoff distances, integrating body and gait cues for robust recognition~\cite{jager2025expanding}. Under such setting, face-only systems achieve under 80\% TAR@1\% FAR and about 84\% Rank-20 accuracy, while multimodal fusion exceeds 90\% TAR, underscoring the need for cues beyond the face for long-range recognition~\cite{liu2024farsight}. \reviewtag{\Rthree}

\noindent \textbf{Cross Age and Child Recognition.}
Age progression causes strong intra-class variation. NIST studies~\cite{grother2019face} show low errors for middle aged adults but higher rates for the youngest and oldest. YFA dataset~\cite{bahmani2022yfa} shows similarity degradation with even short age gaps, for ages under 36 months. Child recognition is reliable only over short term, Systems should use age aware thresholds, periodic re-enrollment, and multimodal fusion. \reviewtag{\Rthree}

\noindent \textbf{Other Specialized Domains.}  
In \textit{mobile authentication and access control}, FR achieves high reliability with emphasis on real time performance and privacy on edge devices~\cite{patel2016continuous}. In \textit{forensic and post mortem identification}, robustness to extreme degradations, aging, and cross spectral imagery remains active, amid debates on accuracy, bias, and evidentiary standards~\cite{zeinstra2018forensic}. These applications illustrate the breadth of FR but merit separate detailed study.  \reviewtag{\Rthree}

\begin{figure}[t]
    \centering
    \includegraphics[width=1.0\linewidth]{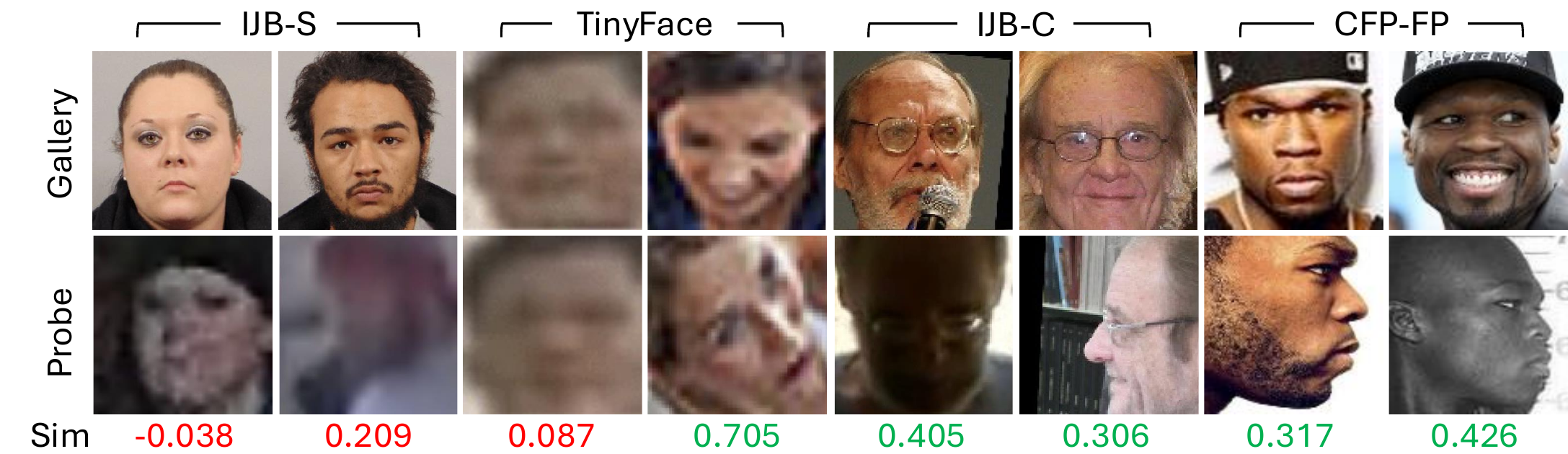}
    \caption{Examples of faces across different datasets: IJB-S~\cite{ijbs}, TinyFace~\cite{cheng2018low}, IJB-C~\cite{ijbc}, and CFP-FP~\cite{cfpfp}. The top row shows gallery images, the middle row shows corresponding probe images, and the bottom row reports cosine similarity scores. Higher similarity scores (in green) indicate successful matches, while lower scores (in red) indicate mismatches. Features are extracted by KP-RPE~\cite{kim2024keypoint} trained on WF4M~\cite{webface}.}
    \label{fig:face_sim}
    \vspace{-4mm}
\end{figure}

\endgroup

\subsection{Multi-modal Recognition: Beyond Facial Imagery}

As FR technologies move towards more challenging environments characterized by low resolution, extreme poses, occlusions, varying illumination conditions, and large-scale databases, reliance solely on facial imagery becomes increasingly insufficient. Real-world scenarios such as surveillance or public safety applications require robust identification techniques capable of handling severely degraded visual information.

To address these challenges, there is growing emphasis on integrating multiple biometric modalities. Incorporating additional cues such as body shape, gait, or even behavioral patterns significantly enhances recognition robustness. Traditionally outputs from multiple biometric modalites are combined using score fusion~\cite{singh2019comprehensive}. 
Score-level fusion combines similarity scores from multiple biometric modalities after similarity comparison. Common approaches include normalization methods like Z-score and min-max, likelihood ratio-based fusion, and simple aggregations such as mean, max, or min fusion~\cite{nandakumar2007likelihood,poh2011unified,poh2007improving,vatsa2007integrating}. These techniques collectively improve robustness and accuracy in challenging recognition scenarios.

On the other hand, multi-modal biometrics can be conducted with the fusion at the input or feature level. SapiensID~\cite{sapiensid} proposes to combine face and body recognition under one model, offering particular promise in cross modality comparison, as body images offer larger visual area that can distinguish individuals at lower image resolutions. The future of robust FR lies in embracing a multi-modal approach, harnessing complementary biometric modalities to overcome the limitations of any single modality.

\begin{table}[t]
\caption{Performance degradation on IJB-S (Survillance to Single protocol) as the gallery size increases with imposters sampled from an external dataset.}
\centering
\resizebox{\linewidth}{!}{
\begin{tabular}{lcccc}
\toprule
\textbf{Gallery Setting} & \textbf{Gallery Size} & \textbf{Rank-1} & \textbf{Rank-5} & \textbf{TPIR @ FPIR=0.01} \\
\midrule
Baseline Gallery      & 202    & 62.0\% & 68.2\% & 46.1\% \\
+1K External Imposters & 1,202  & 56.1\% & 61.6\% & 43.7\% \\
+5K External Imposters & 5,202  & 51.1\% & 57.3\% & 40.8\% \\
+10K External Imposters& 10,202 & 48.4\% & 55.1\% & 38.0\% \\
\bottomrule
\end{tabular}}
\label{tab:ijbs_imposter_gallery}
\vspace{-2mm}
\end{table}

\subsection{Capacity of Generative Models}

An emerging question in synthetic dataset design is not just whether generated faces look realistic, but how many truly distinct and usable identities a generative model can produce. This is fundamentally a question of \textit{identity capacity}: given a fixed number of real training images, how many well-separated subjects can a model generate?

DCFace~\cite{kim2023dcface}, trained on 52k real face images, generates 20k new synthetic identities. In contrast, Vec2Face~\cite{wu2024vec2face}, trained on a much larger dataset (360k images), achieves up to 200k well-separated identities. This scaling behavior demonstrates that generative identity capacity is closely related to the diversity and richness of the real training data.

Recent work by Boddeti {\it et al.}~\cite{boddeti2023biometric} propose a principled statistical framework for estimating the upper bound of this capacity, framing it as a hyperspherical packing problem in the feature space of a FR model. They define capacity as the maximum number of identities that can be placed in this space without exceeding a predefined similarity threshold (related to a false acceptance rate). Their empirical estimates show that StyleGAN3 has a practical upper bound, approximately 1.43 million identities at a 0.1\% FAR, which decreases sharply with stricter thresholds. For class-conditional models like DCFace, the capacity was significantly lower, due to its greater intra-class variation.

\begin{table}[t]
\centering
\caption{Performance Comparison of Foundation Models (FMs) in FR under Different Training Regimes. Accuracies are averaged over LFW, CALFW, CPLFW, CFP-FP, and AgeDB. Rank is 16 for LoRA. CosFace~\cite{wang2018cosface} is used to train the models. }
\label{tab:fm_fr_comparison}
\resizebox{\linewidth}{!}{%
\begin{tabular}{|l|l|l|l|c|}
\hline
\textbf{Model} & \textbf{Arch} & \textbf{Train Dataset} & \textbf{Train Setting} & \textbf{Avg. Acc. (\%)} \\
\hline
DINOv2 & ViT-S & - & Pre-trained (Zero-shot FR) & 64.70 \\
CLIP & ViT-S & - & Pre-trained (Zero-shot FR) & \textbf{82.64} \\
\hline
ViT & ViT-S & 1k IDs & Trained from Scratch & 69.96 \\
DINOv2 & ViT-S & 1k IDs & Fine-tuned (LoRA) & 87.10 \\
CLIP & ViT-S & 1k IDs & Fine-tuned (LoRA) & \textbf{90.75} \\
\hline
ViT & ViT-S & CASIA-WebFace & Trained from Scratch & 88.56 \\
DINOv2 & ViT-S & CASIA-WebFace & Fine-tuned (LoRA) & 90.94 \\
CLIP & ViT-S & CASIA-WebFace & Fine-tuned (LoRA) & \textbf{92.13} \\
\hline
ViT & ViT-L & WebFace4M & Trained from Scratch & 95.65 \\
DINOv2 & ViT-L & WebFace4M & Fine-tuned (LoRA) & \textbf{96.03} \\
CLIP & ViT-L & WebFace4M & Fine-tuned (LoRA) & 95.59 \\
\hline
\end{tabular}%
}
\vspace{-2mm}
\end{table}

These results underscore an important insight: while generative models can amplify identity diversity, their capacity is not unlimited. The sampling distribution remains bounded by the identity entropy encoded during training. Thus, future research can aim to formalize these constraints, explore the theoretical upper bounds of novel identity generation, and propose methods for synthetic identities to be meaningfully distinct and diverse.

This raises a compelling question for the future: could synthetic datasets eventually surpass the utility of real datasets for training FR models? While current synthetic data often lags behind real data due to domain gaps and capacity limitations, the potential advantages of synthetic generation could be unparalleled control over attributes, scalability, and the ability to systematically generate data for rare conditions or underrepresented demographics~\cite{um2024self}. 

Realizing this potential likely requires moving beyond current 2D generative paradigms. Integrating 3D modeling and rendering techniques stands out as a particularly promising direction. By leveraging explicit 3D representations, future generative pipelines could offer physically grounded control over geometry, pose, illumination, and material properties (like skin texture and reflectance), potentially generating synthetic faces with greater realism, diversity, and, crucially, more distinct and well-separated identities than achievable through purely data-driven 2D synthesis alone. DigiFace~\cite{bae2023digiface} explores this direction and the key limitation is in the domain gap. Further research exploring these hybrid approaches, alongside developing better methods to measure and maximize the effective identity capacity, will be key to determining if and how synthetic data can overcome the limitations of, and perhaps even outperform, real-world data collection for advancing FR.

\subsection{Role of Foundation Models in Face Recognition}

Foundation models (FMs) are large-scale models pretrained on extensive image or text datasets for general-purpose tasks, rather than task-specific objectives such as FR. These models provide both pretrained weights and robust feature representations derived from broad visual or textual domains. Chettaoui {\it et al.}~\cite{chettaoui2025froundation} offer a comprehensive overview of the role of foundation models in FR. Their findings indicate that, since FR models are traditionally trained on large-scale datasets, the advantages of using FMs are not clearly observed at the large scale training data.

However, fine-tuning FMs in low-data settings can significantly improve their performance~\cite{chettaoui2025froundation}. Key comparative results are shown in Tab.~\ref{tab:fm_fr_comparison}. However, obtaining large scale training dataset is not difficult for FR, the benefit of FMs is still to be probed. Future work should focus on identifying which fine-tuning techniques, such as LoRA~\cite{hu2022lora}, and which foundation models, like CLIP~\cite{radford2021learning} or DINOv2~\cite{oquab2023dinov2}, offer the best starting points for FR applications. Additionally, there is a need to understand why the advantages of foundation models diminish when training with large-scale FR datasets.

Recently, LAFS~\cite{sun2024lafs} introduces pretraining on unlabeled face data using foundation models, effectively learning critical FR representations and achieving strong few-shot performance. This highlights the value of specialized pretraining and motivates further exploration of domain-specific self-supervised learning (SSL) for developing specialized FR foundation models. It also raises questions about their interaction with general-purpose foundation models and potential reasons why the benefits of general models may diminish on large-scale FR datasets.

\subsection{Interpretability}

\begin{figure}
    \centering
    \includegraphics[width=0.8\linewidth]{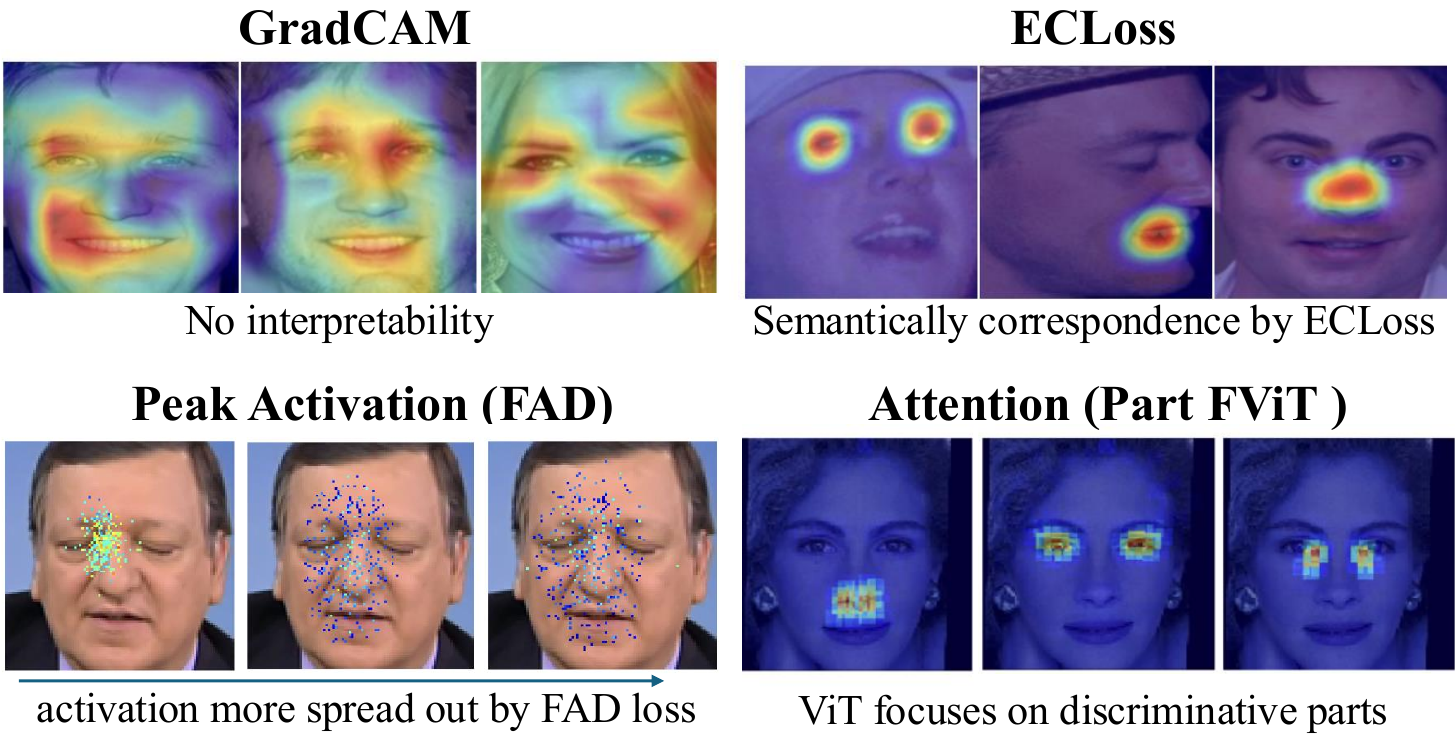}
    \caption{Comparison of visualization methods for face-related tasks. Grad-CAM~\cite{selvaraju2020grad} highlights broad, less interpretable regions, whereas ECLoss~\cite{Lin2024ECLoss} enforces semantic correspondence with activations on meaningful facial parts. Peak Activation with FAD~\cite{yin2019towards} shows that activations become more spread out across the face with the application of FAD loss. Attention maps from PartFViT~\cite{sun2022part} demonstrate that ViT models concentrate on discriminative facial parts, {\it e.g.}, eyes and nose.
}
    \label{fig:activation}
    \vspace{-4mm}
\end{figure}
Deep learning models used for FR are frequently viewed as black boxes. Their internal decision-making processes, involving millions of parameters, are inherently opaque, making it challenging to understand precisely why a particular decision (match vs.~no-match, high vs.~low quality score) was reached. This lack of transparency hinders trust, complicates debugging, and makes it difficult to assess its confidence, and impossible to be presented as evidence in the court. Several interpretability and explainability techniques are being applied or explored in the context of FR:

\noindent \textbf{Saliency/Attribution Maps:} Methods such as Grad-CAM~\cite{selvaraju2020grad} or SHAP~\cite{lundberg2017unified} generate heatmaps highlighting the input image regions (pixels) most influential for the model's decision. For Transformer-based FR models, analyzing internal attention weights can offer insights into which parts of the input representation the model focuses on during processing~\cite{rodis2024multimodal}. FAD~\cite{yin2019towards} proposes spatial activation diversity loss to learn more structured face activation. Some examples are shown in Fig.~\ref{fig:activation}.

\noindent \textbf{Concept-based Explanation:} Moving beyond pixel importance, these approaches aim to link model decisions to higher-level, human-understandable concepts~\cite{xai_concept}. This could involve identifying the influence of specific facial attributes ({\it e.g.}, eye shape, nose structure) or using methods like ECLoss~\cite{Lin2024ECLoss} to directly explain learned features without extra annotations. Some examples are shown in Fig.~\ref{fig:activation}.

\noindent \textbf{Counterfactual Explanation:} These techniques explain a decision by showing minimal changes to the input that would alter the model's output~\cite{stepin2021survey, xai_counterfactual} ({\it e.g.}, ``How would this face need to change to no longer match?''). 

\noindent \textbf{Frequency-Domain Explanation:} Another approach specifically investigated for FR involves analyzing the influence of different frequency components ({\it e.g.}, low vs.~high frequencies representing coarse structure vs.~fine details) in the input images on the matching decision~\cite{Huber2024Frequency, Huber2025FrequencyBias}. This provides a different perspective beyond spatial explanations.

\noindent \textbf{LLM-based Explanation:} 
Recent advances show that large language models (LLMs) are improving the interpretability of FR systems. Traditional tools such as saliency maps or concept attributions highlight key facial regions but fail to provide coherent, human-readable rationales. New approaches like XAI-CLIP~\cite{yao2024xaiclip} and interpretable vision--language alignment methods~\cite{liu2024interpretablevl} use LLMs to generate natural language explanations of why two faces match or differ, citing traits such as ``similar eyebrow curvature, matching nose bridge width, and aligned mouth corners.'' Complementary work on concept bottleneck models~\cite{koh2020concept,dombrowski2023faithful} enhances interpretability by learning high-level concepts (e.g., glasses, facial hair), allowing explanations grounded in semantic features rather than pixel activations. Such language-based reasoning enables FR systems to describe shared and divergent traits transparently, emphasizing the need for tighter grounding between textual justifications and visual evidence.

\begin{figure}
    \centering
    \includegraphics[width=0.75\linewidth]{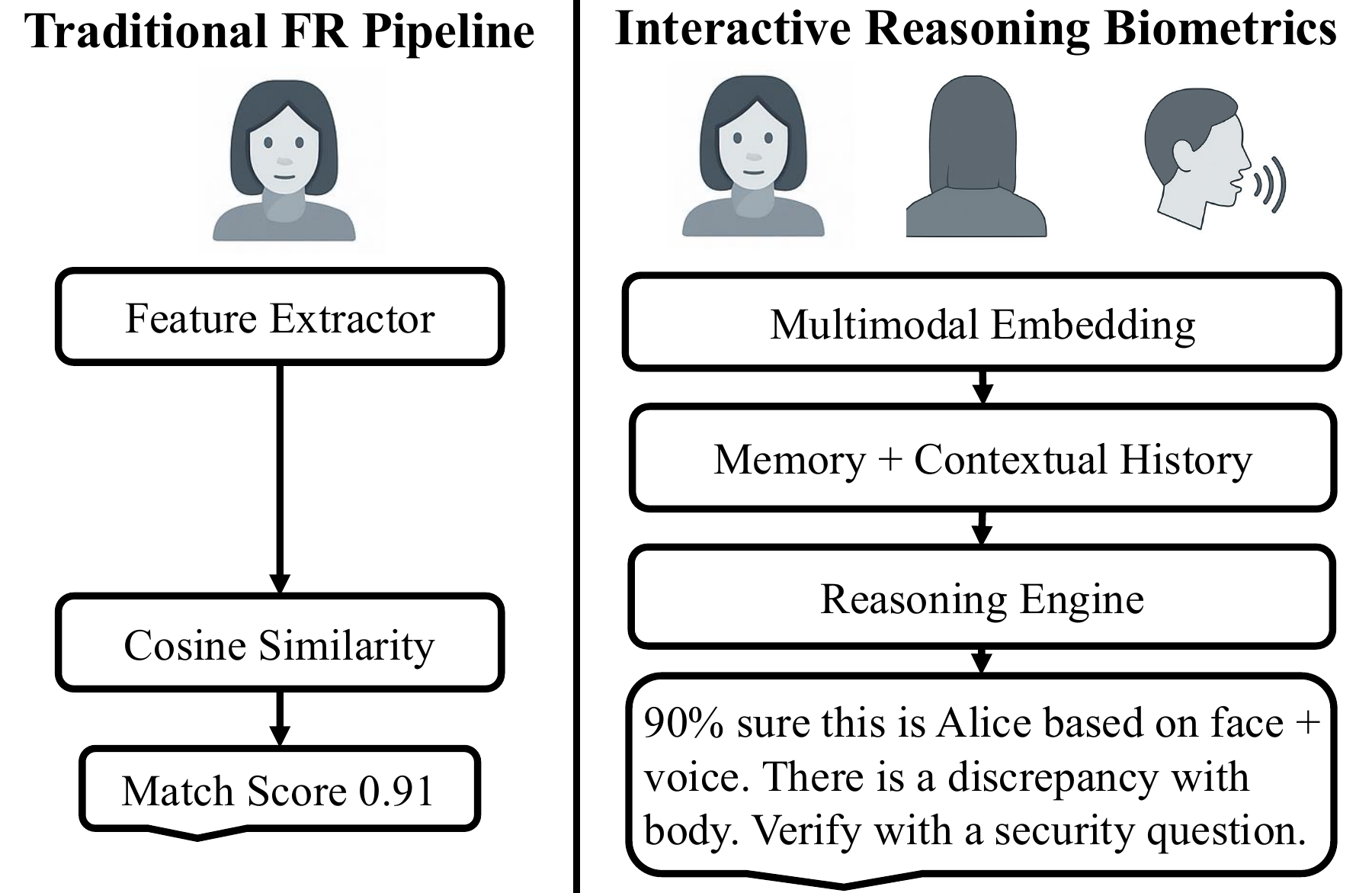}
    \caption{Comparison between traditional FR pipelines and future paradigms integrating multimodal biometrics, reasoning, and explanations. Traditional FR systems output a simple match score based on feature similarity, whereas future systems can reason across modalities, dynamically assess uncertainty, and collaborate with humans through feedback loops.}
    \label{fig:future}
    \vspace{-4mm}
\end{figure}

As interpretability advances, the next challenge is to pair match scores with calibrated confidence and clear reasoning. Future FR systems may, in uncertain cases, simulate multiple identity hypotheses, request additional evidence, or express probabilistic justifications. Fig.~\ref{fig:future} illustrates this shift from static pipelines to interactive, reasoning-based systems that foster human collaboration. A key step toward such feedback loops is enabling models to determine when to involve humans. Building on Face Image Quality Assessment (FIQA) research that estimates input quality or recognizability~\cite{boutros2023cr, ou2024clib, chai2023recognizability}, future work should integrate these signals into broader reasoning frameworks and develop metrics that go beyond accuracy to capture trust, interpretability, and decision quality.






\begingroup
\color{cred}

\subsection{Fairness and Bias in Face Recognition \reviewtag{\Rthree, \Rfour}} \label{sec:fairness}
As FR moved from research to practical deployment, ensuring equitable performance across demographic groups became critical. Fairness concerns intensified in the late 2010s following incidents of wrongful arrests~\cite{hill2020wrongful}, biased airport screening~\cite{garvie2019america}, and misidentification of public figures~\cite{snow2018amazon}. Since 2019, NIST’s FRVT includes demographic analyses~\cite{grother2019face}. For an extensive review, see Kotwal and Marcel~\cite{kotwal2025review}; this section summarizes key sources of bias, datasets, metrics, and mitigation strategies.


\noindent \textbf{Sources of Disparities.}
Bias arises from multiple factors. Training data imbalance has long been cited~\cite{klare2012face,krishnapriya2019understanding}, though disparities persist even in balanced sets~\cite{gwilliam2021rethinking}; Skin reflectance affects accuracy under poor lighting~\cite{cook2019demographic}, while image quality and illumination improvements reduce group gaps~\cite{krishnapriya2020understanding}. Appearance factors such as hairstyle, facial hair, makeup, and occlusion drive gender differences more than gender itself~\cite{kurz2022exploring}.

\noindent \textbf{Datasets and Evaluation.}
Fairness evaluation requires demographically annotated datasets. Public options include RFW~\cite{wang2019racial}, BFW~\cite{robinson2020face}, BUPT and BalancedFace~\cite{wang2018racial}.

\noindent \textbf{Metrics and Mitigation.}
Fairness metrics captures shifts in similarity score distributions. NIST FRVT~\cite{grother2019face} contains large-scale demographic performance variations. Fairness Discrepancy Rate (FDR) quantifies these disparities, with values near one indicating greater parity~\cite{terhorst2020fairness}.
Mitigation occurs through (1) preprocessing via demographic augmentation or synthetic data generation~\cite{kortylewski2019analyzing}, (2) using adaptive architecture, losses or adversarial debiasing~\cite{wang2018racial,mitigating-face-recognition-bias-via-group-adaptive-classifier,gong2020debface}, and (3) \textit{postprocessing} with subgroup calibration or score normalization~\cite{robinson2020face}.

\subsection{Interconnected Areas with FR \reviewtag{\Rthree}}
\label{sec:interconnected}

\noindent\textbf{Face Image Quality Assessment (FIQA).}
Recognition reliability depends strongly on image quality, influenced by pose, blur, and illumination. FIQA models estimate a recognizability score to guide quality aware training and inference. FIQA methods are either unsupervised, predicting quality from recognition feature certainty~\cite{Babnik2022FaceQAN,Babnik2023DifFIQA,meng2021magface,improving-face-recognition-with-a-quality-based-probabilistic-framework}, or supervised, deriving pseudo quality labels through characterization and training an independent model~\cite{boutros2023cr,HernandezOrtega2019FaceQnet,Ou2021SDDFIQA, a-quality-guided-mixture-of-score-fusion-experts-framework-for-human-recognition}.  FIQA now underpins FR by quantifying uncertainty and improving reliability with low quality inputs.

\noindent\textbf{Presentation Attacks and Spoof Detection (PAD).}
As FR systems expand to consumer and border uses, they face attacks like printed photos, replayed videos, and three dimensional masks. PAD defends using texture cues, depth, or multi sensor fusion~\cite{liu2022spoof}. Progress in image forensics and deepfake detection adds augmentation~\cite{Li2020FaceXRay,Li2018WarpingArtifacts}, frequency analysis~\cite{Liu2021SpatialPhase,Luo2021HighFreq}, and disentanglement learning~\cite{Yan2023UCF,Yang2019HeadPose,hierarchical-fine-grained-image-forgery-detection-and-localization}. Interpretable detectors such as DDVQA BLIP~\cite{Zhang2024DDVQA} or M2F2Det~\cite{guo2025rethinking} employ vision language models for explanations, enhancing transparency. 

\noindent\textbf{Privacy Preserving Face Recognition (PFR).}
With FR in sensitive contexts, privacy protection is crucial. Cryptographic techniques use homomorphic encryption or secure multiparty computation for similarity computation without exposing raw faces~\cite{Erkin2009PETS,Yang2023Sensors}, but computationally heavy. Transform based methods conceal by modifying  representations. Early works use obfuscation~\cite{Kevenaar2005AutoID}, while modern ones use adversarial or diffusion methods~\cite{Boutros2023ICCV,Huang2023CVPR}. 

\subsection{Regulatory and Policy Perspectives.}
Global scrutiny has turned fairness and accountability into regulatory imperatives.
The \textit{GDPR}~\cite{eu_gdpr_2016,zaborska2019biometric} classifies facial imagery and other biometric data as special personal data, requiring explicit consent or anonymization of major FR datasets.
The \textit{EU AI Act}~\cite{eu_ai_act_2024} designates biometric identification as high risk, mandating transparency, bias evaluation, and human oversight.
Meanwhile, cities like San Francisco and Boston restrict public FR use, while others expand it for security purposes.
These measures embed fairness as both scientific and regulatory.
\reviewtag{\Rthree, \Rfour}

\endgroup

\section{Summary}
\label{sec:conclusion}

Over the past fifty years, FR has advanced from geometric and handcrafted methods to deep learning models surpassing human accuracy. Progress in architectures (ResNets, ViTs), loss functions (margin based softmax), and large datasets ({\it e.g.}, WebFace42M) has driven state of the art performance. Despite success on benchmarks (LFW~\cite{lfw}, CFP FP~\cite{cfpfp}, IJB-C~\cite{ijbc}) and deployment in authentication and security, challenges persist: scalability to billions, degraded imagery, limited interpretability, data privacy and fairness.

Future work should pursue multimodal and explainable systems, realistic synthetic data, and foundation models, emphasizing large scale robustness, confidence calibration, and ethical use through consent, security, and purpose limitation.
\textcolor{cred}{Recent developments of FR datasets in the community have highlighted growing attention to the ethical and legal implications of datasets that lack clear and informed user consent, with increasing efforts to promote transparency and responsible data sourcing. \reviewtag{\Rone, \Rthree}}

\quad\\ 
\noindent\textbf{Acknowledgments} This research is based upon work supported by the Office of the Director of National Intelligence (ODNI), Intelligence Advanced Research Projects Activity (IARPA), via 2022-21102100004. The views and conclusions contained herein are those of
the authors and should not be interpreted as necessarily representing the official policies, either expressed or implied, of ODNI, IARPA, or the U.S. Government. The U.S. Government is authorized to reproduce and distribute reprints for governmental purposes notwithstanding any copyright annotation therein.

\bibliographystyle{IEEEtran}
\bibliography{bib}

\vspace{-10mm}
\begin{IEEEbiography}
[{\includegraphics[width=1in,height=1.25in,clip,keepaspectratio]{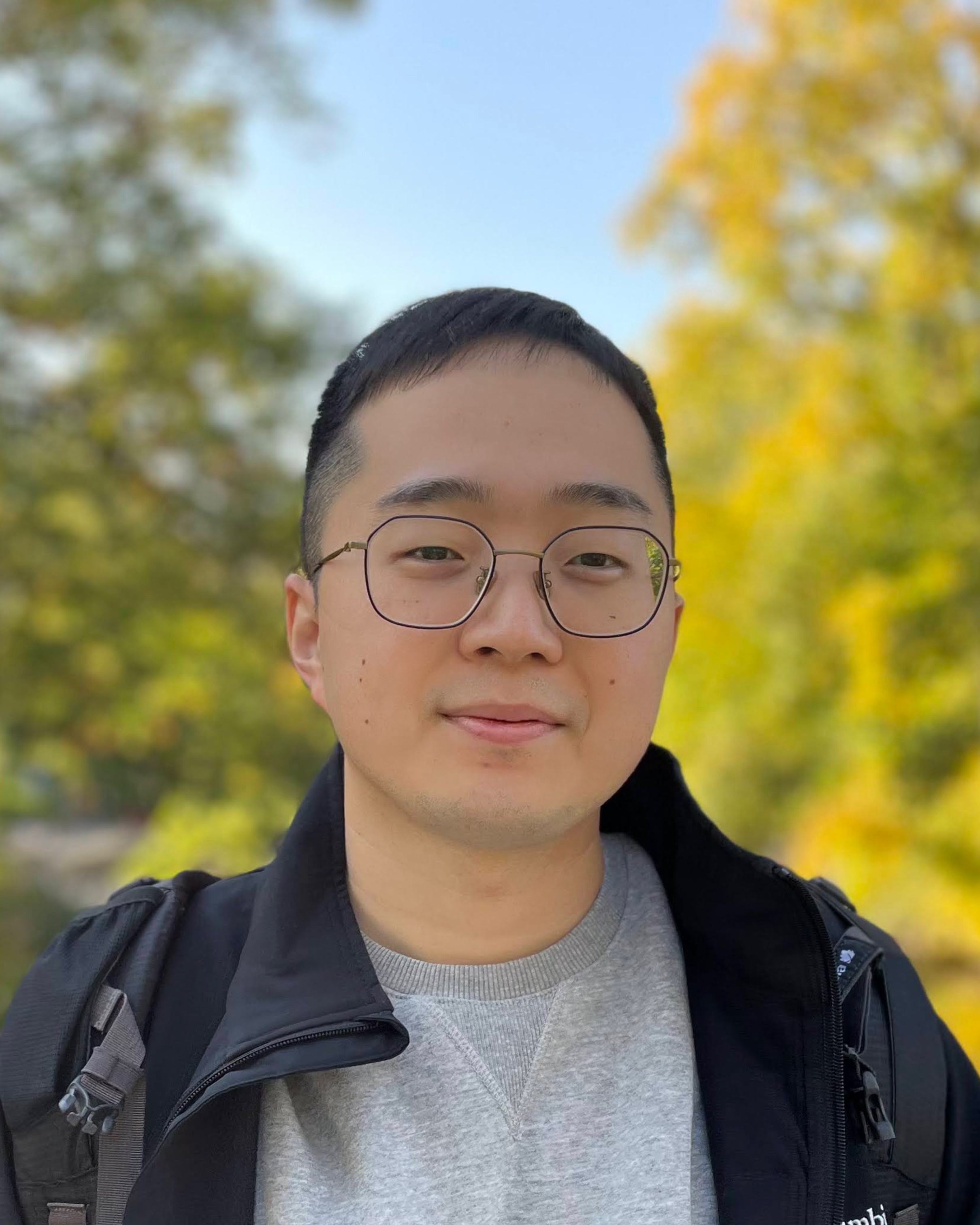}}]{Minchul Kim}
received his Ph.D. degree in Computer Science and Engineering from Michigan State University, East Lansing, MI, USA. He is currently a software engineer at Google, where he works on machine learning and computer vision applications. His research interests include face recognition, biometrics, and deep learning. During his Ph.D., he published in top-tier conferences and journals in the field of computer vision and biometrics.
\end{IEEEbiography}

\vspace{-10mm}
\begin{IEEEbiography}
[{\includegraphics[width=1in,height=1.25in,clip,keepaspectratio]{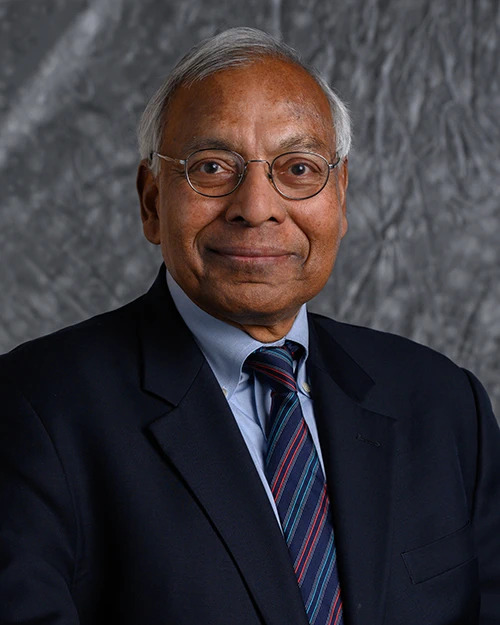}}]{Anil K. Jain}
(\IEEEmembership{Life Fellow,~IEEE}) is a University Distinguished Professor in the Department of Computer Science and Engineering at Michigan State University. His research interests include pattern recognition, computer vision, and biometric authentication. 
Jain is a member of the U.S. National Academy of Engineering, the Indian National Academy of Engineering, the World Academy of Sciences, and the Chinese Academy of Sciences.
\end{IEEEbiography}

\vspace{-10mm}
\begin{IEEEbiography}
[{\includegraphics[width=1in,height=1.25in,clip,keepaspectratio]{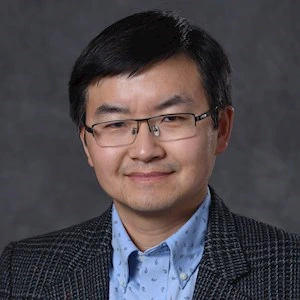}}]{Xiaoming Liu}
(\IEEEmembership{Fellow,~IEEE}) is a MSU Foundation Professor, and Anil and Nandita Jain Endowed Professor in the Department of Computer Science and Engineering at Michigan State University. He received his Ph.D.~from Carnegie Mellon University in 2004. His research interests span computer vision, machine learning, and biometrics.  
He is an Associate Editor for  IEEE Transactions on Pattern Analysis and Machine Intelligence. He is a fellow of IEEE and IAPR.
\end{IEEEbiography}

\vspace{\fill}

\end{document}